\documentclass[lettersize,journal]{IEEEtran}
\usepackage{amsmath,amsfonts}
\usepackage{algorithmic}
\usepackage{algorithm}
\usepackage{array}
\usepackage[caption=false,font=normalsize,labelfont=sf,textfont=sf]{subfig}
\usepackage{textcomp}
\usepackage{stfloats}
\usepackage{url}
\usepackage{verbatim}
\usepackage{graphicx}
\usepackage{cite}
\usepackage{booktabs} 
\usepackage{multirow} 
\usepackage{siunitx}  
\usepackage{subcaption} 
\usepackage{adjustbox} 

\begin{document}

\title{State Space Models over Directed Graphs}


\author{Junzhi She, Xunkai Li, Rong-Hua Li, Guoren Wang

\thanks{This paper was produced by the IEEE Publication Technology Group. They are in Piscataway, NJ.}
\thanks{Manuscript received month xx, 20xx; revised month xx, 20xx.}}

\markboth{IEEE TRANSACTIONS ON BIG DATA,~Vol.~xx, No.~x, month1/month2~20xx}%
{Shell \MakeLowercase{\textit{et al.}}: A Sample Article Using IEEEtran.cls for IEEE Journals}

\IEEEpubid{\makebox[\textwidth]{\parbox{\dimexpr\textwidth-2em\relax}{\centering 2332-7790~\copyright~2025 IEEE. Personal use is permitted, but republication/redistribution requires IEEE permission. \\ See \url{https://www.ieee.org/publications/rights/index.html} for more information.}}}

\maketitle

\begin{abstract}  
Directed graphs are ubiquitous across numerous domains, where the directionality of edges encodes critical causal dependencies. However, existing GNNs and graph Transformers tailored for directed graphs face two major challenges: (1) effectively capturing long-range causal dependencies derived from directed edges; (2) balancing accuracy and training efficiency when processing large-scale graph datasets.  
In recent years, state space models (SSMs) have achieved substantial progress in causal sequence tasks, and their variants designed for graphs have demonstrated state-of-the-art  accuracy while maintaining high efficiency across various graph learning benchmarks. However, existing graph state space models are exclusively designed for undirected graphs, which limits their performance in directed graph learning.  To this end,
we propose an innovative approach DirEgo2Token which sequentializes directed graphs via k-hop ego graphs. This marks the first systematic extension of state space models to the field of directed graph learning. Building upon this, we develop DirGraphSSM, a novel directed graph neural network architecture that implements state space models on directed graphs via the message-passing mechanism. Experimental results demonstrate that DirGraphSSM achieves state-of-the-art performance on three representative directed graph learning tasks while attaining competitive performance on two additional tasks with 1.5× to 2× training speed improvements compared to existing state-of-the-art models.  
\end{abstract}


\section{Introduction}

A wide range of complex real-world systems—from digital circuits and neural networks to software code and citation networks—can be naturally modeled as directed graphs. Unlike undirected graphs, crucial information in these systems is often encapsulated in long-range causal dependencies determined by the directionality of edges.  For instance, in the field of Electronic Design Automation (EDA), the critical path delay spanning multiple logic levels dictates whether a chip meets its timing constraints \cite{EDA_OCB, EDA_HLS}. In deep learning, the update of model parameters relies on gradients that are backpropagated throughout the network, which is crucial for model performance \cite{NA_search, DVAE}. Similarly, in program analysis, specific execution paths can be used to determine if code exhibits malicious behavior \cite{Malware_detect}; and in citation networks, the influence of a seminal work propagates through citation chains, shaping the development of an entire research field \cite{DAGFormer, citation_pubmed}. Therefore, effectively capturing and leveraging these long-range causal dependencies is not only a central theoretical challenge in directed graph learning but also a key bottleneck in numerous high-impact applications \cite{ogbg-code2_dataset, DAGNN, DAGFormer}.

Current mainstream approaches for directed graph learning primarily fall into two categories: Directed GNNs and Directed Graph Transformers. Directed GNNs adapt the message-passing paradigm, employing either spatial methods (e.g., DirGNN \cite{dirgnn} and D-HYPR \cite{D-HYPR}) or spectral methods (e.g., MagNet \cite{magnet} and HoloNets \cite{HoloNets}) to encode directionality. While these methods outperform standard GNNs, their reliance on local aggregation makes them susceptible to the well-known over-squashing problem (i.e., information loss in deep graph neural networks \cite{oversquash}) and over-smoothing problem (i.e., node features becoming indistinguishable \cite{oversmooth}). Consequently, their ability to model long-range dependencies in large-scale graphs is severely limited.
On the other hand, directed graph transformers leverage a global attention mechanism and incorporate directionality through directed  positional encodings, such as magnetic Laplacian and directed random walk positional encodings \cite{DirGraphTransformer, DirPE}. Their global receptive field allows them to overcome the limitations of GNNs, demonstrating superior performance on tasks requiring long-range reasoning. However, the quadratic computational complexity of both the global attention mechanism and directed positional encodings makes them prohibitively expensive for large-scale graphs, restricting their practical application.

The trade-off between the efficiency of GNNs and the effectiveness of Transformers has spurred new research into alternative architectures. Sparse graph Transformers are representative methods. For example, GraphGPS replaces dense attention with linear attention layers originating from natural language processing \cite{GraphGPS}, while Exphormer leverages graph  augmentation techniques to heuristically adjust the attention receptive field \cite{Exphormer}. These methods achieve a certain trade-off between efficiency and accuracy but still have limitations: on one hand, linear attention was originally designed for sequential data, and its accuracy significantly decreases when applied to irregular graph structures; on the other hand, heuristic graph augmentation may distort the original topological structure, weakening long-range causal dependencies. Moreover, these methods are designed for undirected graphs. When applied to directed graphs, they require expensive directed positional encodings to achieve high accuracy, thereby introducing new performance bottlenecks.

\IEEEpubidadjcol

Recently, State Space Models (SSMs) have emerged as a powerful alternative, demonstrating linear time complexity and exceptional long-range dependency modeling capabilities for sequential data \cite{S4D, Mamba, Mamba2}. This success has inspired their application to graph data, giving rise to models like GraphMamba \cite{GraphMamba} and GMN (Graph Mamba Networks) \cite{GMN}. These Mamba-based graph SSMs (\textbf{graph Mamba models}) sequentialize graph structures by ordering nodes based on centrality metrics, showing superior performance and efficiency compared to sparse graph transformers. However, a critical issue has been overlooked: these methods are designed for undirected graphs and utilize the sequential scanning and batch processing mechanisms of Mamba \cite{Mamba} , thus failing to address the unique challenges posed by directed graphs and their causal dependencies.

Specifically, applying existing graph Mamba models to directed graphs exposes several fundamental limitations:
\begin{enumerate}
    \item \textbf{Loss of Causal Dependencies}: Existing sequentialization  strategies in graph Mamba models are typically direction-agnostic. This oversight prevents the models from capturing the long-range causal paths that define information flow in directed systems.
    \item \textbf{Arbitrary Node Ordering}: existing graph Mamba models rely on a deterministic node ordering. However, the complex topology of directed graphs means no single ordering criterion can represent the graph's information flow without loss. This inherent ambiguity compromises the model's robustness and permutation invariance. (For instance, a PageRank-based ordering yields a significantly different node sequence than a degree-based one; with a degree-based ordering, nodes with identical out-degrees produce a random permutation).
    \item \textbf{Computational Inefficiency from Sequentialization}: Flattening a topologically imbalanced directed graph into a one-dimensional sequence results in highly non-uniform output lengths. This necessitates extensive padding during batch training, leading to significant computational and memory overhead, especially for large-scale heterogeneous directed graphs.
    \item \textbf{Serial computation  Dependency}: The selective scanning mechanism in graph Mamba models introduces RNN-like serial dependencies in computation, lowering its efficiency \cite{Mamba}.
\end{enumerate}
Consequently, current graph Mamba models  struggle to simultaneously meet the accuracy and efficiency requirements for large-scale directed graph learning tasks.

To address these challenges, we propose the \textbf{Directed Graph State Space Model} (DirGraphSSM), a novel architecture specifically designed for directed graph learning. Our main contributions are as follows:
\begin{itemize}
    \item \textbf{Permutation-Invariant DiGraph Sequentialization}. We propose a novel framework called \verb|DirEgo2Token|  , which makes SSMs permutation-invariant on directed graphs. For each node, this framework constructs its directed k-hop ego-network and groups neighbor nodes by their shortest path distance, thereby generating a canonical sequence. This method effectively encodes long range directed topology into a sequence that can be processed by an SSM while remaining invariant to the initial node ordering.
    \item  \textbf{Efficient Parallelizable Architecture}. We develop an efficient message-passing-based implementation for DirGraphSSM. This design enables parallel computation over a node's neighborhood, eliminating the serial computation dependency  and  the costly padding operation during batch processing in the graph Mamba models, thereby significantly accelerating training and inference.
    \item  \textbf{State-of-the-Art Efficiency and Performance}.  We conduct extensive experiments that demonstrate the superior performance of DirGraphSSM. On large-scale directed graph benchmarks, our model trains 1.5 to 2 times faster than state-of-the-art graph Mamba models while matching or surpassing their accuracy.
\end{itemize}

The remainder of this paper is organized as follows: Section \ref{section:related_work} reviews related work on directed graph learning, sequence modeling, and graph sequence models; Section 3 introduces the necessary preliminaries; Section 4 elaborates on our proposed DirGraphSSM model; Section 5 reports the comprehensive experimental results and analysis; finally, Section 6 concludes this paper.

\section{RELATED WORK}
\label{section:related_work}

\subsection{Directed Graph Learning} 
Directed graphs have garnered significant attention from researchers in graph learning due to their capacity to model directionality and potential causal relationships inherent in real-world networks. The primary approaches can be broadly categorized into two families: spectral-domain and spatial-domain GNNs for directed graphs. The core idea of spectral methods is to construct a suitable Laplacian operator for non-symmetric directed graphs, mapping node features into the spectral domain and enabling efficient convolution through polynomial approximations, such as Chebyshev polynomials \cite{GCN}. Representative works include \textbf{MagNet}~\cite{magnet}, which leverages the magnetic Laplacian to explicitly encode directionality using complex phases, and \textbf{DiGCN}~\cite{Digcn}, which constructs an operator based on the stationary distribution of PageRank and employs sparse approximation with personalized PageRank. Spatial methods, on the other hand, directly extend the message-passing paradigm by differentiating the aggregation or parameterization for in-coming and out-going neighborhoods. For instance, \textbf{DGCN}~\cite{DGCN} performs parallel convolutions on first-order and two types of second-order proximities. \textbf{DirGNN}~\cite{dirgnn} conducts separate message aggregation for in- and out-neighborhoods followed by a joint update. \textbf{ADPA}~\cite{ADPA} utilizes various directed motifs to drive training-free propagation and employs a two-level attention mechanism for adaptive aggregation. Despite their different starting points, both lines of work fundamentally rely on information aggregation within local neighborhoods, making them susceptible to over-smoothing and struggling to capture long-range dependencies. To address this, recent directed graph Transformers attempt to aggregate global information while preserving directionality. Examples include positional encodings based on the phase of the magnetic Laplacian~\cite{DirGraphTransformer} and reachability positional encodings derived from directed random walks or personalized PageRank~\cite{DirGraphTransformer}. These methods can inject directional information without altering the attention architecture but often incur high computational costs. \textbf{DAGFormer}~\cite{DAGFormer}, in contrast, leverages the partial order of DAGs to contract the attention receptive field and designs a lightweight positional encoding, achieving a better balance between accuracy and efficiency. These advancements have laid the groundwork for subsequent methods that integrate global modeling with direction-aware mechanisms. 

\subsection{Sequence Modeling} 

Sequence modeling has evolved from probabilistic graphical models to attention mechanisms, and more recently, to continuous state-space modeling. Early approaches like \textbf{Hidden Markov Models (HMMs)}~\cite{HMM} described sequences using finite state transitions and emission probabilities but were limited by the Markov assumption, making it difficult to capture long-term dependencies. Subsequently, \textbf{Recurrent Neural Networks (RNNs)}~\cite{RNN} introduced recurrent hidden states, and gated units such as \textbf{LSTM}~\cite{lstm} and \textbf{GRU}~\cite{GRU} mitigated the vanishing gradient problem. However, their sequential computation paradigm introduced a significant parallelization bottleneck. The \textbf{Transformer}~\cite{transformer} replaced recurrence with self-attention, enabling direct modeling of relationships between any two positions and achieving substantial efficiency gains through parallel training. The subsequent development of \textbf{BERT}~\cite{bert} and \textbf{GPT}~\cite{GPT} established the pre-training paradigm. Nevertheless, the quadratic complexity and memory consumption of self-attention became a bottleneck for extremely long sequences. In recent years, \textbf{State Space Models (SSMs)} have emerged, viewing sequences as continuous dynamical systems. They achieve near-linear complexity through linear state updates driven by input, which facilitates streaming and long-sequence inference. Representative methods such as \textbf{S4}~\cite{S4} and \textbf{S4D}~\cite{S4D} achieve efficient training and strong expressive power through structured state matrices, stable discretization, and fast convolutions. \textbf{Mamba}~\cite{Mamba, Mamba2} further introduces selective parameterization, allowing state updates to adapt dynamically to the input. This endows the model with both content-awareness and long-term memory, achieving performance comparable or superior to Transformers of similar scale on multi-modal and long-sequence tasks. Compared to attention, SSMs exhibit favorable properties such as cache/scan-friendliness, stable gradients, and hardware parallelism. Furthermore, their equivalence to 1D convolutions and the frequency response of rational functions provides a theoretical foundation for understanding and improving these models, such as balancing state dimensionality versus channel count and coordinating stability with expressiveness. These properties establish SSMs as an efficient primitive for long-sequence modeling, capable of replacing or complementing attention mechanisms. 

\subsection{Graph Sequence Models} Making  graph data sequential for processing by powerful sequence models, thereby balancing graph structure modeling with training efficiency, has become a significant research direction at the intersection of graph learning and sequence modeling. Directly flattening an entire graph into a sequence encounters the issue of node ordering, where the number of possible permutations is $A_{n}^{2}$, making it infeasible for sequence models to process exhaustively. Therefore, researchers have developed more refined sequentialization strategies by leveraging graph structure.  
\textbf{DeepWalk} \cite{DeepWalk} and \textbf{Node2Vec} \cite{Node2Vec} use random walks to sample node sequences, treating them as "sentences" within the graph. These methods naturally incorporate graph structure into sequences through the process of random walking.
Focusing on Directed Acyclic Graphs (DAGs), \textbf{PACE}~\cite{PACE} addresses the "injection" problem. Its \verb|dag2seq|  component first  canonicalizes the graph to obtain a unique node ordering, then injects predecessor topology via structured positional encodings. This ensures that non-isomorphic graphs are not mapped to the same sequence, and a masked Transformer is used to capture the partial order. With the rise of SSMs, \textbf{GraphMamba}~\cite{GraphMamba} flattens the entire graph using a simple node priority sorting (based on centrality metrics), replaces the global attention branch with Mamba, and leverages the selective mechanism to capture global dependencies in linear time. It also enhances robustness by permuting the order of nodes with the same priority during training. \textbf{Graph Mamba Networks (GMN)}~\cite{GMN} uses random walks to sample multi-scale subgraphs as tokens, which are sorted by priority and receptive field length, and processed with a bidirectional Mamba.  These works demonstrate that research on graph sequentialization continues to optimize the preservation of structural information and processing efficiency. However, it is worth noting that sequentialization introduces trade-offs, such as order sensitivity and token budget limitations (e.g., the impact of sorting on permutation invariance, and the allocation ratio between global and local tokens). Further trade-offs among order sensitivity, expressive power, and computational efficiency remain to be addressed.

\section{PRELIMINARIES}

\subsection{digraph representation learning}
We consider a directed graph (digraph) $\mathcal{G} = (\mathcal{V}, \mathcal{E})$ consisting of $|\mathcal{V}| = n$ nodes and $|\mathcal{E}| = m$ directed edges. Each node is associated with a feature vector of dimension $f$, collectively represented by the feature matrix $\mathbf{X} \in \mathbb{R}^{n \times f}$. The structure of $\mathcal{G}$ is encoded by an asymmetric adjacency matrix $\mathbf{A} \in \{0,1\}^{n \times n}$, where $\mathbf{A}(u, v) = 1$ if and only if $(u, v) \in \mathcal{E}$, and $0$ otherwise. Depending on the task, the graph may be accompanied by different types of labels: for node classification, each node has a one-hot label over $c$ classes, represented by $\mathbf{Y} \in \mathbb{R}^{n \times c}$; for graph classification, each graph is assigned a one-hot label of size $c$, with labels aggregated in $\mathbf{Y} \in \mathbb{R}^{n \times c}$; and for graph regression, each graph is associated with a real-valued scalar, stored in $\mathbf{Y}\in \mathbb{R} ^1$. The goal of directed graph representation learning is to learn a mapping function that leverages both the directed graph topology and node features to produce node-level or graph-level embeddings, which can then be used for label prediction in downstream tasks.

\subsection{Directed Message Passing}
\label{section:preliminary:mpnn}

In Graph Neural Networks (GNNs), the message passing mechanism serves as a key approach to extend convolutional operations to graph-structured data. Its core idea is to update the representation of a target node by aggregating information from its neighboring nodes (source nodes).

The critical distinction between directed and undirected message passing lies in the definition of neighbors and the direction of message flow. In undirected graphs, edges lack directionality; the neighbor set typically includes all adjacent nodes, and message passing is bidirectional and symmetric. In contrast, directed graphs have explicit edge directions, where messages propagate only from source to target nodes (or vice versa, depending on the chosen directionality), and the aggregation process depends on topological relationships defined by either incoming or outgoing edges. This distinction enables directed GNNs to better model directional semantics such as information flow, dependencies, and causality, making them more suitable for real-world scenarios with widespread directed structural data.

Specifically, to leverage both incoming and outgoing directional information, directed GNNs often employ separate parameters for message passing along the in-directions and out-directions, and then combine the two to produce the final result \cite{dirgnn}. More concretely, a bidirectional directed message passing mechanism defines distinct message functions and aggregation processes for in-edges and out-edges. Let $\mathcal{N}_{in}(i)$ denote the set of in-neighbors of node $i$ (i.e., nodes pointing directly to $i$), and $\mathcal{N}_{out}(i)$ denote the set of out-neighbors of node $i$ . The aggregated message from the in-direction $\mathbf{m}_{i,\text{in}}^{(k)}$ and the out-direction $\mathbf{m}_{i,\text{out}}^{(k)}$ can be computed respectively as:
\begin{equation}
    \begin{aligned}
\mathbf{m}_{i,\mathrm{in}}^{(k)} &= \underset{j\in \mathcal{N} _{in}(i)}{\mathrm{AGG}_{\mathrm{in}}}\left( \phi _{\mathrm{in}}^{(k)}\left( \mathbf{x}_{i}^{(k-1)},\mathbf{x}_{j}^{(k-1)},\mathbf{e}_{j,i} \right) \right) ,
\\
\mathbf{m}_{i,\mathrm{out}}^{(k)} &= \underset{j\in \mathcal{N} _{out}(i)}{\mathrm{AGG}_{\mathrm{out}}}\left( \phi _{\mathrm{out}}^{(k)}\left( \mathbf{x}_{i}^{(k-1)},\mathbf{x}_{j}^{(k-1)},\mathbf{e}_{i,j} \right) \right) , \label{eq:dir-mpnn-agg}
    \end{aligned}
\end{equation}
where $\mathbf{x}_{i}^{(k-1)} \in \mathbb{R}^{f}$ denotes the node features of node $i$ at layer $(k-1)$, and $\mathbf{e}_{j,i} \in \mathbb{R}^{d}$ denotes the (optional) edge features from node $j$ to node $i$; $\phi_{\text{in}}^{(k)}$ and $\phi_{\text{out}}^{(k)}$ are two distinct learnable functions responsible for generating messages over in-edges and out-edges, respectively; $\text{AGG}_{\text{in}}$ and $\text{AGG}_{\text{out}}$ are the corresponding aggregation functions (e.g., sum, mean, etc.). Finally, the updated node representation is obtained by combining the two directional messages with the node’s previous features via a combination function $\psi^{(k)}$ \cite{dirgnn}:
\begin{equation}
\mathbf{x}_i^{(k)} = \text{COM}^{(k)} \left( \mathbf{x}_i^{(k-1)}, \psi^{(k)} \left( \mathbf{m}_{i,\text{in}}^{(k)}, \mathbf{m}_{i,\text{out}}^{(k)} \right) \right), \label{eq:dir-mpnn-com}
\end{equation}
where $\psi^{(k)}$ may be a concatenation, weighted summation, or other effective combination operation. This approach explicitly distinguishes the two relational directions in a directed graph, allowing more flexible modeling of node dependencies under different semantic contexts.
Note that bidirectional message passing is not mandatory in directed GNNs. In some cases, message passing may be applied only along in-edges or out-edges.

\subsection{State Space Models}
\label{preliminary:ssm}

State Space Models (SSMs) are a classic class of sequence models that map a 1D input sequence $x(t) \in \mathbb{R}$ to an output sequence $y(t) \in \mathbb{R}$ through a latent state $h(t) \in \mathbb{R}^{D}$. This process is described by a linear ordinary differential equation (ODE) \cite{first_ssm}:
\begin{equation}
\begin{aligned}
h^{\prime}(t)&=\boldsymbol{A}\,h(t)+\boldsymbol{B}\,x(t),
\\
y(t)&=\boldsymbol{C}\,h(t)
\end{aligned}\label{eq:ssm-linear-differential}
\end{equation}
where $\boldsymbol{A} \in \mathbb{R}^{D \times D}$ is the state matrix, and $\boldsymbol{B} \in \mathbb{R}^{D \times 1}$ and $\boldsymbol{C} \in \mathbb{R}^{1 \times D}$ are the input and output matrices, respectively. $D$ represents the hidden dimension. Since continuous systems are difficult to handle directly in deep learning contexts, they typically need to be discretized. By introducing a discretization step size $\Delta$, the continuous system above can be transformed into a discrete form \cite{discrete_ssm}:
\begin{equation}
\begin{aligned}
	h_t&=\overline{\boldsymbol{A}}\,h_{t-1}+\overline{\boldsymbol{B}}\,x_t,\\
	y_t&=\overline{\boldsymbol{C}}h_t,\\
\end{aligned}\label{eq:ssm-discrete-form}
\end{equation}
Here, the discretized parameters $\overline{\boldsymbol{A}}$ and $\overline{\boldsymbol{B}}$ are obtained using the Zero-Order Hold (ZOH) method, specified as $\overline{\boldsymbol{A}}=\exp \left( \Delta \boldsymbol{A} \right)$ and $\overline{\boldsymbol{B}}=\left( \Delta \boldsymbol{A} \right) ^{-1}\left( \exp \left( \Delta \boldsymbol{A}-\boldsymbol{I} \right) \right) \cdot \Delta \boldsymbol{B}$. This discrete form reveals the recurrent or recursive nature of SSMs, enabling them to process sequences step-by-step, much like Recurrent Neural Networks (RNNs).

Furthermore, by unrolling the recursive process, the SSM can be shown to be equivalent to a convolution operation. The output sequence $y$ can be expressed as the convolution of the input sequence $x$ with a structured convolution kernel $\overline{\boldsymbol{K}}$ \cite{convolution_discrete_ssm}:

\begin{equation}
\begin{aligned}
	y_k&=\overline{\boldsymbol{C}}\overline{\boldsymbol{A}}^{\mathrm{k}}\overline{\boldsymbol{B}}x_0+\overline{\boldsymbol{C}}\overline{\boldsymbol{A}}^{\mathrm{k}-1}\overline{\boldsymbol{B}}x_1+\cdots +\overline{\boldsymbol{C}}\overline{\boldsymbol{B}}x_k\\
	\overline{\boldsymbol{K}}&=\left( \overline{\boldsymbol{C}}\overline{\boldsymbol{B}},\overline{\boldsymbol{C}}\overline{\boldsymbol{A}}\overline{\boldsymbol{B}},...,\overline{\boldsymbol{C}}\overline{\boldsymbol{A}}^{L-1}\overline{\boldsymbol{B}} \right)\\
	\boldsymbol{y}&=\boldsymbol{x}*\overline{\boldsymbol{K}}\\
\end{aligned}\label{eq:ssm-convolution-form}
\end{equation}
Here, $\overline{\boldsymbol{K}} \in \mathbb{R}^{L}$ is known as the SSM convolution kernel, where $L$ is the length of the input sequence. This convolutional form allows the model to be computed in parallel during training, significantly improving efficiency. Building on this, Structured State Space sequence model (S4) significantly enhance the efficiency and scalability of SSMs through reparameterization techniques \cite{S4}, emerging as a powerful alternative to the attention mechanism.
However, directly computing the convolution kernel $\overline{\boldsymbol{K}}$ of length $L$ has a time  complexity of $O(D^2L)$, which becomes a bottleneck when processing long sequences. To improve the efficiency of kernel computation, Gu et al.\cite{S4D} proposed S4D, which significantly accelerates the computation by using a diagonalized matrix $\boldsymbol{A}$. Specifically, when $\boldsymbol{A}$ is a diagonal matrix, the computation of the SSM convolution kernel can be simplified by leveraging the properties of Vandermonde matrices, resulting in a nearly linear time complexity of $O(DL)$.

Despite the significant efficiency breakthroughs of structured SSMs, their parameters ($\boldsymbol{A}, \boldsymbol{B}, \boldsymbol{C}$) are time-invariant, meaning the model's information processing mechanism is fixed and lacks the ability to adapt dynamically to the context. To address this limitation, a new generation of SSMs, Mamba, introduced a selection mechanism \cite{Mamba}. Its core idea is to make key SSM parameters functions of the input data. In this way, Mamba can dynamically modulate the state transitions and output projections based on the current input, thereby selectively allowing relevant information to flow through while ignoring irrelevant information, effectively achieving context-aware information compression.

Building on the theoretical foundation of Mamba, Gu et al.\cite{Mamba2} further proposed the SSD framework, which reveals a deep connection between SSMs and Transformers with causal masking through structured matrices. The linear recurrence of an SSM is equivalent to the quadratic computation of a Structured Masked Attention (SMA). This insight provides a crucial foundation for our subsequent design of a unified model architecture that integrates SSMs and attention.

\section{METHODOLOGY}

\subsection{Overall Architecture}

\begin{figure*}[!t] 
    \centering
    \includegraphics[width=0.8\textwidth]{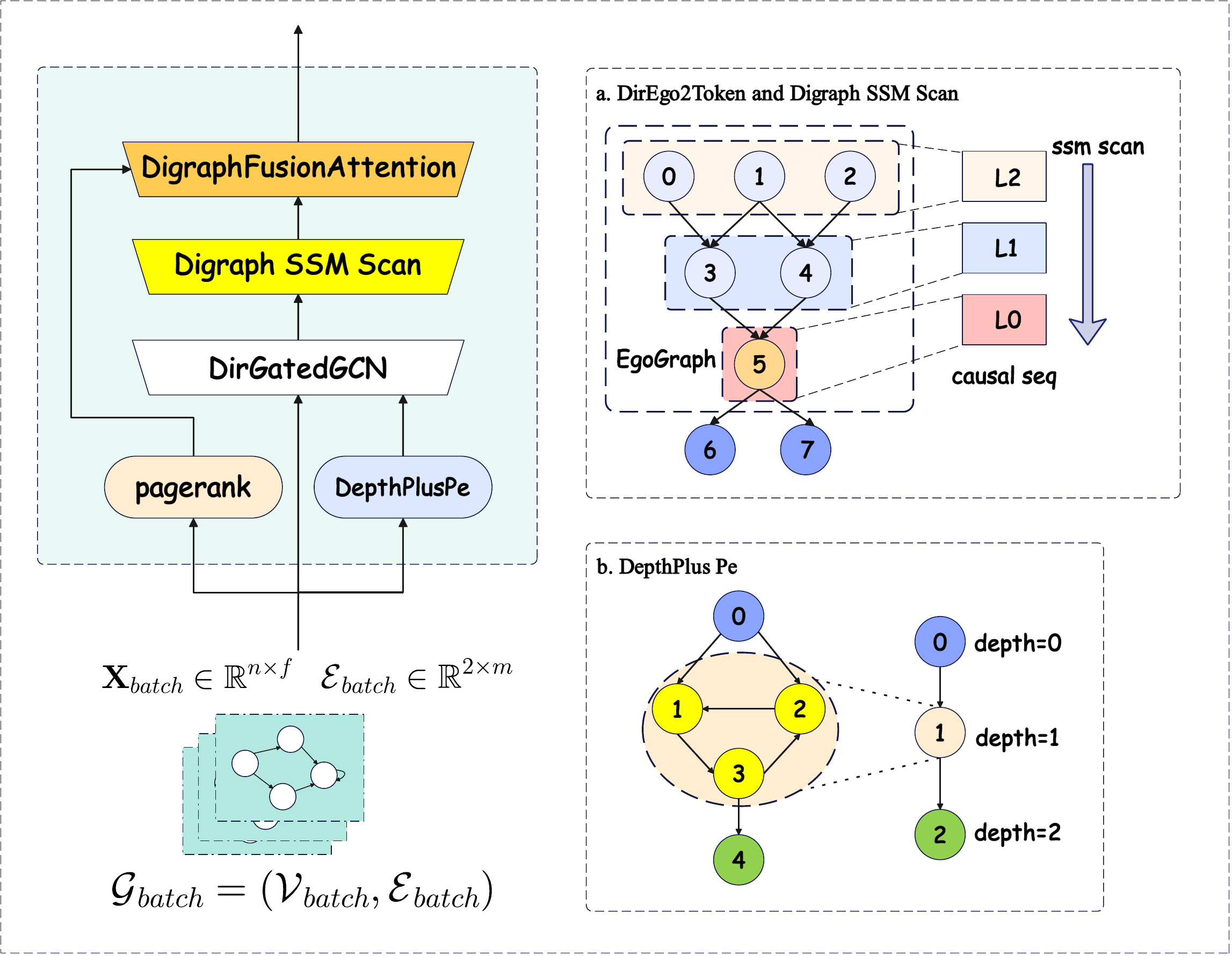} 
    \caption{Overall Architecture of DirGraphSSM. DirGraphSSM first employs DepthPlus to generate positional encodings that reflect the hierarchical structure of nodes, which are then integrated into the node features. Subsequently, DirGatedGCN is applied to produce enhanced node representations. This is followed by a SSM scan on directed ego graph of each node. Finally, the Digraph Fusion Attention module leverages the PageRank to guide the feature fusion across different attention spaces, resulting in the updated node representations.}
    \label{fig:overallArchitecture}
\end{figure*}

\label{method:Architecture}
This chapter details the architecture of our proposed DirGraphSSM model, as illustrated in Figure \ref{fig:overallArchitecture}. The core idea is to apply State Space Models (SSMs) to directed graph learning, leveraging their strengths in capturing long-range directed dependencies while maintaining computational efficiency.

To capture the long-range causal dependencies in directed graphs using SSMs, we first introduce a novel directed graph sequentialization framework, \textbf{DirEgo2Token}, which lays the foundation for capturing long-range causal relationships within directed graphs (see Section \ref{method:DirEgo2Token} for details).
Building upon this sequentialization concept, we construct the core SSM module \textbf{Digraph SSM Scan}, implemented in an efficient message-passing paradigm with multi-head attention as a selective mechanism (see Section \ref{method:digraphSSMScan} for details).

Furthermore, to capture richer topological information and enhance the model's ability to represent causal dependencies, 
we design \textbf{DirGatedGCN Structural Encoding} and \textbf{DepthPlus Positional Encoding}.
(see Section \ref{method:se_pe} for details).
Subsequently, to effectively integrate node representations across multiple attention heads, we devise a \textbf{Digraph Fusion Attention} module to further augment the model's representation capabilities (see Section \ref{section:DigraphFusionAttention} for details).

Finally, we provide a complexity analysis of DirGraphSSM, demonstrating its efficiency advantages in processing large batches of sparse directed graphs (see Section \ref{section:method:ComplexityAnalysis} for details).

\subsection{DiGraph Sequentialization: DirEgo2Token}
\label{method:DirEgo2Token}
A robust graph sequentialization method must encode the structural information of the graph into a sequence while being resilient to node permutations. For directed graphs, the precedence relationships among nodes  derived from the edge directionality are particularly crucial. Existing graph sequentialization methods are often designed for undirected graphs or fail to establish a canonical ordering for certain nodes, making them unsuitable for directed graphs. Therefore, we propose DirEgo2Token, a novel digraph sequentialization method. Its core idea is to construct a sequence for each central node that reflects its "causal history," yielding a representation that is inherently invariant to node permutation. Figure \ref{fig:overallArchitecture}.a intuitively illustrates the overall flow of this module.

Specifically, inspired by NAGphormer's   ego-graph based sequentialization of undirected graphs \cite{NAGphormer, NAGphormer+}, we utilize a directed ego-graph to construct a causal sequence for each node $v$. For a given directed graph $G=(V, E)$ and a central node $v \in V$, the construction process of its corresponding causal sequence $S_v$ can be formally described as follows:

\begin{equation}
\begin{aligned}
	&\mathcal{N} _{in}^{k}(v)=\{u\in V\mid \mathrm{SPD(}u,v)\le k\}\\
	&L_i=\{u\in \mathcal{N} _{in}^{k}(v)\mid \mathrm{SPD(}u,v)=i\},i\in \left[ 0,k \right]\\
	&S_v=(L_k,L_{k-1},...,L_1,L_0)\\
\end{aligned}\label{eq:DirEgo2Token}
\end{equation}
Here, $\mathcal{N} _{in}^{k}(v)$ denotes the set of $k$-hop directed predecessors of node $v$. This set contains all nodes that can reach $v$ via a directed path with a shortest path distance (SPD) of at most $k$. $\text{SPD}(u, v)$ represents the shortest path distance between nodes $u$ and $v$, and $L_0 = \{v\}$.

Overall, the aforementioned sequentialization process can be viewed as a hierarchical directed information diffusion or causal tracing process. Information "flows" step-by-step from distant predecessors to the central node, naturally encoding the directed dependencies.

\subsection{Digraph SSM Scan}
\label{method:digraphSSMScan}
Based on the causal sequences generated by \verb|DirEgo2Token| (Section \ref{method:DirEgo2Token}), we propose the \textbf{Digraph SSM Scan} method , which efficiently models the causal dependencies using a State Space Model (SSM). 

Given a central node $v$ and its corresponding causal sequence $S_v = (L_k, L_{k-1}, \dots, L_0)$, DirGraphSSM first aggregates the node set of each hop $L_i$ into a "causal signal" vector. It then scans this sequence of signals using an SSM to ultimately generate a representation $\mathbf{y}_v$ for the central node $v$. Specifically, this process is formalized as:
\begin{equation}
    \begin{aligned}
	\mathbf{z}_i&=\mathrm{Aggregate(}L_i,v), i\in \left[ 0,k \right]\\
	\mathbf{y}_v&=\mathrm{SSM}_{\mathrm{conv}}(\mathbf{z}_k,\mathbf{z}_{k-1},\dots ,\mathbf{z}_0)\\
\end{aligned}
\end{equation}
where $\mathbf{z}_i$ is the aggregated vector for the node set of the $i$-th hop, and $\mathrm{SSM}_{\text{conv}}$ denotes the SSM scan process implemented in a convolutional form.

To effectively capture the dependency strength of the nodes in $L_i$ on the central node $v$, we employ a Multi-head Attention mechanism to implement the $\mathrm{Aggregate}(L_i, v)$ operation. Specifically, for the node set of each hop $L_i$, we compute its attention-weighted sum with respect to the central node $v$:
\begin{equation}
    \mathrm{Aggregate(}L_i,v)=\sum_{u\in L_i}{\frac{\kappa (\mathbf{x}_v,\mathbf{x}_u)}{\sum_{w\in \mathcal{N} _{in}^{k}(v)}{\kappa (\mathbf{x}_v,\mathbf{x}_w)}}}\cdot f(\mathbf{x}_u)
    \label{eq:digraphssm-Aggregate}
\end{equation}
where $\kappa(\cdot, \cdot)$ is the attention kernel function, defined as:
\begin{equation}
\kappa (\mathbf{x}_v,\mathbf{x}_u)=\exp \left( \frac{\langle f\left( \mathbf{x}_v \right) \mathbf{W}_Q,f\left( \mathbf{x}_u \right) \mathbf{W}_K\rangle}{\sqrt{d_k}} \right)
\label{eq:digraphssm-attn-weight}
\end{equation}
Here, $f(\mathbf{x}_v)$ is the feature vector of node $v$ after incorporating structural and positional encodings (Section \ref{method:se_pe}) and undergoing a projection transformation.
Notably, this attention mechanism plays a role in  SSM scan process analogous to the "selective scanning" in Mamba \cite{Mamba}. It can dynamically adjust attention weights based on the input, and the entire computation process is independent of the sequence order, allowing for full parallelization.

Furthermore, by leveraging the distributive and associative laws of matrix operations, the sequence scanning process can be equivalently expressed in a message passing paradigm:

\begin{equation}
\begin{aligned}
	\mathbf{y}_v&=\sum_{u\in \mathcal{N} _{in}^{k}(v)}{\alpha _{u,v}}\cdot \mathrm{SSM}^{(\mathrm{spd(}u,v))}(f\left( \mathbf{x}_u \right) \mathbf{W}_V)\\
	\alpha _{u,v}&=\frac{\kappa (\mathbf{x}_v,\mathbf{x}_u)}{\sum_{w\in \mathcal{N} _{in}^{k}(v)}{\kappa (\mathbf{x}_v,\mathbf{x}_w)}}\\
\end{aligned} \label{eq:digraphssm-mpnn}
\end{equation}
Here,  $\alpha_{u,v}$ is the attention weight from node $u$ to $v$ , $\mathrm{SSM}^{(k)} = \overline{\boldsymbol{C}}\overline{\boldsymbol{A}}^{k}\overline{\boldsymbol{B}}$, with $\mathbf{A} \in \mathbb{R}^{D \times D}$, $\mathbf{C} \in \mathbb{R}^{d \times D}$, and $\mathbf{B} \in \mathbb{R}^{D \times d}$. $d$ denotes the total hidden state dimension corresponding to the attention mechanism (i.e., the original dimension before dividing by the number of heads $h$), and $D$ is the internal hidden state dimension of the SSM. (Unlike the scalar convolution kernel in Section \ref{preliminary:ssm}, we linearly expand the input signal dimension from 1 to $d$.)

This message passing interpretation shows that  \verb|Digraph SSM Scan|  is essentially an attention-weighted multi-hop information aggregation mechanism. Each message is transformed by an SSM convolution kernel corresponding to the path distance before being passed to the central node.
Thanks to this message passing formulation, our implementation avoids explicitly constructing sequences, thus eliminating the issues of inconsistent sequence lengths caused by different graph structures and the associated padding overhead. Specifically, during the preprocessing stage, we compute the $k$-hop predecessor set for each central node and record their shortest path distances ($\mathrm{spd}$) using Breadth-First Search. This information is used to construct \verb|k_hop_edge_index| and \verb|k_hop_spd|, which enables the parallel SSM scan to be implemented within a message passing framework during training.

\subsection{Structural and Positional Encodings}
\label{method:se_pe}

To  better capture the local and global topological structures of directed graphs, we introduce a local structural encoding based on DirGatedGCN and a global positional encoding based on the depth of nodes in directed graphs.

\textbf{DirGatedGCN Structural Encoding:} Enhancing initial node representations using the powerful local structure extraction capabilities of GNNs can effectively improve the expressive power of Graph Transformers \cite{GraphGPS, SAT}. Inspired by this, we use a directed GatedGCN \cite{gatedGCN}, extended from the DirGNN framework \cite{dirgnn}, to generate local structural encodings for nodes \ref{section:preliminary:mpnn}. This approach models neighborhood information from incoming and outgoing edges with distinct parameters.

GatedGCN enhances traditional graph convolutions by introducing a gating mechanism, thereby improving the model's adaptability and expressive power:
\begin{equation}
    \begin{aligned}
	\boldsymbol{h}_{v}^{l}&=\boldsymbol{h}_{v}^{l-1}\boldsymbol{W}_{1}^{l}+\sum_{u\in \mathcal{N} (v)}{\boldsymbol{\eta }_{v,u}}\odot \boldsymbol{h}_{u}^{l-1}\boldsymbol{W}_{2}^{l}\\
	\boldsymbol{\eta }_{v,u}&=\sigma (\boldsymbol{h}_{v}^{l-1}\boldsymbol{W}_{3}^{l}+\boldsymbol{h}_{u}^{l-1}\boldsymbol{W}_{4}^{l})\\
\end{aligned} \label{eq:GatedGCN}
\end{equation}
where $\sigma$ is the Sigmoid activation function, $\boldsymbol{W}_{1}^{l}$, $\boldsymbol{W}_{2}^{l}$, $\boldsymbol{W}_{3}^{l}$, and $\boldsymbol{W}_{4}^{l}$ are layer-$l$-specific trainable weight matrices, $\mathcal{N}(v)$ is the set of neighbors of node $v$, and $\odot$ denotes the element-wise product (Hadamard product).
This structural encoding strategy provides the subsequent SSM module with node features rich in local structural prior knowledge, achieving a complementary relationship between the local perception capability of GNNs and the global modeling capability of SSMs.

\textbf{DepthPlus Positional Encoding :}
In directed graphs, especially directed acyclic graphs (DAGs), nodes possess hierarchical information, termed depth, which is determined by the direction of edges. Incorporating this information as positional encoding into a Graph Transformer can effectively enhance the model's expressive power \cite{DAGFormer}. The depth of a node $v$ in a DAG is defined as follows:
\begin{equation}
depth(v)=\begin{cases}
	0&		degree_{in}\left( v \right) =0\\
	1+\max_{(u,v)\in E} depth(u)&		\mathrm{otherwise}\\
\end{cases} \label{eq:depth}
\end{equation}
The $depth(v)$ can be computed using a topological sort algorithm on the DAG. This concept is analogous to the sequential ordering of tokens in a text sequence and can be used to generate positional encodings in a sinusoidal form \cite{DAGFormer}:
\begin{equation}
\begin{aligned}
PE_{(v,2i)} &= \sin\left(\frac{\mathrm{depth}(v)}{10000^{\frac{2i}{d}}}\right) \\
PE_{(v,2i+1)} &= \cos\left(\frac{\mathrm{depth}(v)}{10000^{\frac{2i}{d}}}\right)
\end{aligned} \label{SinusoidalPE}
\end{equation}
While many real-world directed graphs exhibit clear hierarchical structures, the presence of cycles renders topological sorting inapplicable for computing $depth(v)$. To provide meaningful global hierarchical positional encodings for nodes in directed cyclic graphs without violating the ordering determined by directed edges, we propose the \textit{DepthPlus} method. This method generalizes the concept of "depth" from DAGs to arbitrary directed graphs. Figure \ref{fig:overallArchitecture}.b intuitively illustrates the core idea of this method.

The steps of the DepthPlus method are as follows: 
First, we use Tarjan's algorithm \cite{Tarjan-algorithm} to decompose the original directed graph $G=(V, E)$ into a set of Strongly Connected Components (SCCs). Each SCC is a set of nodes from the original graph, which we denote as $\{C_1, C_2, \ldots, C_k\}$.
Next, we construct an acyclic condensation graph $G'=(V', E')$, where $V' = \{c'_1, c'_2, \ldots, c'_k\}$ is a set of \textbf{supernodes}, and each supernode $c'_i$ uniquely corresponds to an SCC node set $C_i$. An edge exists from supernode $c'_i$ to $c'_j$ in $G'$ if and only if there exist nodes $u \in C_i$ and $v \in C_j$ in the original graph $G$ such that $(u,v) \in E$.
Since $G'$ is a DAG, we can compute the depth of each supernode $c'_i$, denoted as $depth'(c'_i)$, using a topological sort algorithm. Finally, we assign the depth value of a supernode to all the original nodes within its corresponding SCC. That is, for any original node $v$, if $v \in C_i$, its final depth $depth(v)$ is defined as:
\begin{equation}
    depth(v) = depth'(c'_i) \quad \text{where } v \in C_i \label{eq:DepthPlus}
\end{equation}
In this way, even in directed cyclic graph, each node can be assigned an effective depth value that reflects its position in the macroscopic hierarchical structure.

By combining the structural and positional encodings described above, we obtain the enhanced representation $f(\mathbf{x}_v)$ for each node $v$, which is then used for the subsequent SSM scan.

\begin{equation}
\begin{aligned}
	\mathbf{p}_{v}^{(\mathrm{pos)}}&=DepthPlus(v)\\
	f(\mathbf{x}_v)&=DirGatedGCN(\mathbf{x}_v+\mathbf{p}_{v}^{(\mathrm{pos)}})\\
\end{aligned} \label{eq:se-and-pe}
\end{equation}

\subsection{Digraph Fusion Attention}
\label{section:DigraphFusionAttention}

\begin{figure*}[!t] 
    \centering
    \includegraphics[width=1.0\textwidth]{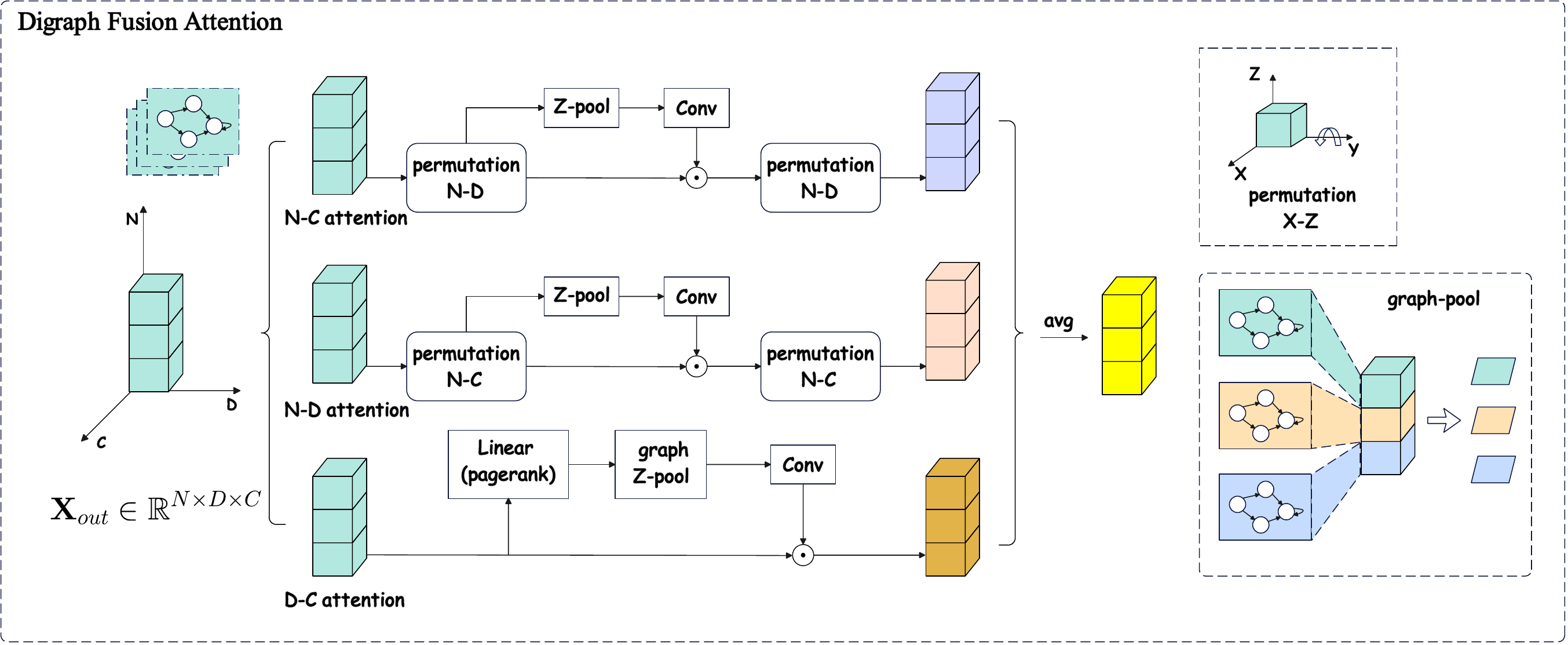} 
    \caption{Digraph Fusion Attention. It integrates multi-channel node features through three parallel branches that capture cross-dimensional interactions between nodes (N), features (D), and channels (C). It incorporates PageRank-based structural priors to guide feature aggregation over dimension N and uses permutation and pooling operations to capture interaction between dimension N-C and N-D. The outputs of the three branches are averaged to produce enhanced node representations.}
    \label{fig:DigraphFusionAttention}
\end{figure*}

To integrate node features from different attention heads and enhance the model’s representational ability, we draw inspiration from the triplet attention mechanism in computer vision \cite{TripletAttention} and design a directed graph fusion attention module (Digraph Fusion Attention). This module treats the representations learned by different attention heads as distinct channels and captures cross-dimensional dependencies across the number of nodes (N), feature dimensions (D), and the number of channels (C) through three parallel branches. Furthermore, it utilizes PageRank, a prior structural feature unique to directed graphs, to guide feature aggregation along the graph dimension, enabling effective interaction of information across different attention heads. The framework of this module is shown in Figure \ref{fig:DigraphFusionAttention}.

PageRank is a classic algorithm for ranking the importance of nodes in a directed graph \cite{PageRank}. Its core idea is that a node is more important if it is pointed to by other important nodes. This importance score, $PR(u)$, can be computed iteratively:
\begin{equation} \label{eq:pagerank-cal}
PR(u)=\frac{1-a}{N}+a\sum_{v\in N_{in}(u)}{\frac{PR(v)}{degree_{out}(v)}}
\end{equation}
where $N_{in}(u)$ is the set of 1 hop incoming neighbors of node $u$, $degree_{out}(v)$ is the out-degree of node $v$, $a$ is the damping factor, and $N$ is the total number of nodes in the graph. As this value naturally reflects structural importance in a directed graph, it is used as a prior weight to guide feature fusion.

Building upon the foundational concepts outlined above, we now proceed to a formal exposition of the computational logic underlying this module.
For an input tensor $X \in \mathbb{R}^{N \times D \times C}$, its first, second, and third dimensions correspond to N, D, and C, respectively. To flexibly aggregate information across different dimensions, we first introduce a permutation operation $P$. Let $P_{k \to 1}$ denote the permutation operation that moves the $k$-th dimension of a tensor to the first dimension, and $P^{-1}_{k \to 1}$ be its inverse. Next, we define the \textbf{Z-Pool} operation, which concatenates the max-pooled and average-pooled statistics of an input tensor along its first dimension:
\begin{equation} \label{eq:ZPool}
\mathrm{ZPool}(Y)=[\mathrm{MaxPool}_1(Y),\mathrm{AvgPool}_1(Y)]
\end{equation}
where $[\cdot, \cdot]$ denotes the concatenation operation. By first applying a permutation to the input tensor and then applying Z-Pool, we can achieve information compression along any arbitrary dimension. Similarly, we define \textbf{graph Z-Pool} ($\mathcal{G} \mathrm{Pool}$) to aggregate node features within each graph of a batch:
\begin{equation}
\begin{aligned} \label{eq:graphZPool}
\mathcal{G} \mathrm{Pool}(Y,\mathrm{batch}) = [ & \mathrm{GMaxPool}(Y,\mathrm{batch}), \\
& \mathrm{GAvgPool}(Y,\mathrm{batch}) ]
\end{aligned}
\end{equation}
where $\mathrm{GMaxPool}$ and $\mathrm{GAvgPool}$ are graph-level max and average pooling operations, respectively, performed by default on the first dimension.

Based on the above definitions, the core computation of this module can be summarized by the following equations, which consist of three parallel attention branches and a final fusing step:
\begin{equation} \label{eq:multi-channel-feature-merge}
\begin{aligned}
	{X^{\prime}}_C&=P_{C\rightarrow 1}(X),{X^{\prime}}_D=P_{D\rightarrow 1}(X)\\
	W_{ND}&=\sigma (\mathrm{Conv}1\mathrm{d(ZPool(}{X^{\prime}}_C)))\\
	W_{NC}&=\sigma (\mathrm{Conv}1\mathrm{d(ZPool(}{X^{\prime}}_D)))\\
	w_p&=\mathrm{Soft}\max\mathrm{(Linear(}p))\\
	W_{DC}&=\mathrm{Broadcast(}\sigma (\mathrm{Conv}2\mathrm{d(}\mathcal{G} \mathrm{Pool(}X\otimes w_p,\mathrm{batch))))}\\
	X_{out}&=\frac{1}{3}\bigl( P_{C\rightarrow 1}^{-1}({X^{\prime}}_C\otimes W_{ND})+ \\
	&\quad P_{D\rightarrow 1}^{-1}({X^{\prime}}_D\otimes W_{NC}) + X\otimes W_{DC} \bigr)
\end{aligned}
\end{equation}
where $\sigma$ is the Sigmoid activation function and $\otimes$ represents element-wise multiplication (with broadcasting support).

In the \textbf{N-D and N-C interaction} branches, to capture dependencies between different dimensions, we first perform a permutation. For instance, to capture the interaction between the N and D dimensions, we permute the channel dimension C to the first position, apply the Z-Pool operation to compress information along this dimension, and then generate the attention weights $W_{ND}$ through a 1D convolutional layer. These weights are applied to the permuted tensor $X'_{C}$, and finally, the inverse permutation operation is used to restore its original dimensional order. The N-C interaction process is analogous, with the target for permutation and compression being the feature dimension D.

In the \textbf{D-C interaction} branch, we first compute the structural importance weights $w_p$ using the nodes' PageRank values $p$. Subsequently, the  $\mathcal{G} \mathrm{Pool}$  operation is employed to aggregate graph-level statistics, which uses the batch index to ensure that no information is leaked between different graphs within the batch. The resulting graph-level features are passed through a 2D convolutional layer and a graph-level broadcast operation to generate the attention weights $W_{DC}$, which are then applied to each node.

Finally, the outputs from the three branches are fused via an averaging operation to obtain the final enhanced node features $X_{out}$. Our DirGraphSSM model employs a multi-head attention mechanism, so the number of channels C in the feature fusing module corresponds to the number of attention heads. This module further strengthens the information interaction between different attention heads with minimal computational overhead, thereby effectively improving the model's representation capability.

\subsection{Bidirectional Scan}
\label{section: Bidirectional Scan}

To comprehensively capture the dependencies in directed graphs, we introduce an optional bidirectional scan mechanism. This mechanism is designed to simultaneously model the information flow originating from both predecessor nodes (forward direction) and successor nodes (reverse direction). It draws inspiration from the concepts of Bidirectional LSTMs \cite{bilstm} and Directed Graph Neural Networks (DirGNN) \cite{dirgnn}.

In terms of implementation, we first construct the reverse graph $G_{rev}$ by inverting the direction of all edges in the original graph $G$. Subsequently, the model independently applies the \verb|Digraph SSM Scan| module to both the original graph $G$ and the reverse graph $G_{rev}$. Structural features such as the causal sequence, positional encodings, and PageRank are also computed independently for the reverse graph. Finally, the node representations obtained from the forward and reverse scans are fused (e.g., through concatenation and projection) to yield a more complete node representation.

The bidirectional scan mechanism can significantly expand the model's effective receptive field. When stacking multiple layers, this mechanism enables the construction of information pathways between nodes that are unreachable via unidirectional paths. For instance, if nodes $v_1$ and $v_2$ share a common predecessor $u$, information can be passed from $v_1$ to $u$ via a reverse scan in one layer, and subsequently from $u$ to $v_2$ via a forward scan in the next layer. This pattern allows the model to capture more complex topological dependencies, thereby effectively enhancing its overall representational power in certain scenarios.

\subsection{Computational Complexity Analysis}
\label{section:method:ComplexityAnalysis}

This section analyzes the computational complexity of the DirGraphSSM model. The analysis is divided into 2 main parts: a one-time preprocessing step for the entire dataset  and the forward propagation process for each batch of graphs. We use the following notation: $n = |V|$ is the number of nodes, $m = |E|$ is the number of edges, $d$ is the hidden state dimension of the attention module, $K$ is the maximum number of hops, $L=K+1$ is the maximum sequence length, $p_k$ is the average size of the $k$-hop predecessor set for a node, $D$ is the internal state dimension of the SSM, and $C$ is the number of multi-head attention heads. For simplicity, we set the hidden dimensions of various linear projection layers and the initial node feature dimension to $d$, without affecting the asymptotic complexity analysis.

\subsubsection{Preprocessing Complexity}
The preprocessing stage is performed once on the entire dataset before training begins, and its cost can be amortized. It primarily includes:
\begin{itemize}
    \item  \textbf{DepthPlus Positional Encoding}: This process first uses Tarjan's algorithm \cite{Tarjan-algorithm} to find all strongly connected components (SCCs) in the graph, which has a time complexity of $O(n+m)$. Next, a condensation graph is constructed, and a topological sort is performed on it, which also takes $O(n+m)$ time. Therefore, the total complexity of \verb|DepthPlus| is $O(n+m)$.
    \item \textbf{PageRank Computation}: The standard PageRank algorithm is computed iteratively. For a fixed number of iterations $I$ (typically a small constant), its complexity is $O(I \cdot (n+m))$, which simplifies to $O(n+m)$.
\end{itemize}
In summary, the computational complexity of the entire one-time preprocessing stage is dominated by graph traversal algorithms, resulting in a complexity of $O(n+m)$.

\subsubsection{Forward Propagation Complexity}
Forward propagation constitutes the main computational cost during both training and inference, and it consists of the following 4 parts: 
\begin{itemize}
    \item \textbf{SSM Kernel Precomputation}: Before forward propagation begins, the SSM convolution kernel needs to be precomputed. The complexity of this process depends on the maximum sequence length $L = k + 1$, the internal SSM state dimension $D$, and the model hidden dimension $d$. We adopt the parameterization method of the structured state space model S4D \cite{S4D}, enabling this process to be efficiently completed with a complexity of $O(L \cdot D \cdot d)$. This computational cost is independent of the graph size ($n$ and $m$), so its overhead becomes negligible when processing large-scale graphs.
    \item \textbf{Structural Encoding (DirGatedGCN)}: We use DirGatedGCN to generate local structural encodings for the nodes. For a single DirGatedGCN layer, its computational complexity is primarily determined by message aggregation over all edges and feature transformation for all nodes, resulting in a total complexity of $O(m \cdot d^2 + n \cdot d^2) = O((n+m)d^2)$.
    \item \textbf{Digraph SSM Scan}:  Its computational complexity depends on the total number of $k$-hop predecessor connections for all nodes, denoted as $|E_k| = \sum_{v \in V} |\mathcal{N}_{in}^{k}(v)| = n \cdot p_k$. For each message from a predecessor node $u$ to a central node $v$, the computational cost mainly comes from the calculation of multi-head attention scores and the SSM transformation, with the former having a complexity of $O(d^2)$ and the latter $O(d)$. Therefore, the total complexity of the DiGraphSSM module is $O(|E_k| \cdot d^2) = O(n \cdot p_k \cdot d^2)$. For sparse directed graphs where the $K$-hop neighborhood is much smaller than the total number of nodes (i.e., $p_K \ll n$), the algorithm exhibits linear complexity, leading to strong scalability and significant efficiency advantages.
    \item \textbf{Digraph Fusion Attention}: This module operates on a tensor of size $n \times d \times C$. The complexity of its internal pooling, lightweight convolution, and element-wise operations is linear with respect to the number of nodes $n$. Thus, the total complexity of this module is $O(n \cdot d \cdot C)$.
\end{itemize}

In summary, the total complexity of the model's forward propagation is approximately $O((n + m)d^2 + n \cdot p_k \cdot d^2)$. By cleverly transforming the SSM scanning process into a message-passing paradigm, our model successfully links the computational complexity to the size of the $k$-hop directed neighborhood $p_k$ rather than the square of the total number of nodes $n$. This avoids the $O(n^2)$ computational bottleneck inherent in the global graph attention mechanism used in standard graph Transformers, making DirGraphSSM highly efficient and scalable when processing large-scale sparse directed graphs.

\newpage

\section{EXPERIMENTS}

In this section, we systematically evaluate the performance of DirGraphSSM in terms of both accuracy and efficiency through a series of experiments. Firstly, we introduce the experimental setup. Then, we report the performance results of DirGraphSSM on both DAG and directed cyclic graph datasets compared to several representative baseline models, and further conduct a comparative analysis of the computational efficiency across different models. Finally, we investigate the intrinsic characteristics and effectiveness of the proposed method via parameter sensitivity analysis and ablation studies.

\subsection{Experimental Setup}

\subsubsection{Datasets}

\begin{table*}[htbp]
\centering
\caption{Dataset statistic.}
\label{tab:dataset-statistic}
\begin{tabular}{cccccc}
\toprule
\multicolumn{1}{c}{Statistic} & \multicolumn{1}{c}{ogbg-code2} & \multicolumn{1}{c}{MalNet-Tiny}  & \multicolumn{1}{c}{NA} & \multicolumn{1}{c}{EDA-HLS}  &  \multicolumn{1}{c}{self-citation} \\
\midrule
Num Graphs & 452,741 & 5,000 & 19,020 & 18,570 & 1,000  \\
Avg Nodes per Graph & 125.2 & 1,410.3 & 8 & 94.7 & 59.1  \\
Min Nodes per Graph & 11 & 5 & 8  & 4 & 30  \\
Max Nodes per Graph & 36,123 & 5,000 & 8 & 474 & 296  \\
Avg Edges per Graph & 124 & 2,859.9 & 11.5 & 122.12 & 123.3  \\
\midrule
Total Nodes & 56,683,173 & 7,051,500 & 152,160 & 1,758,579 & 59,100  \\
Avg $p_{\infty}$ per Node &  4.89  & 8.77 & 3.5 & 25.9 & 2.15 \\
Total $p_{\infty}$ & 277,180,717 & 61,871,976 & 532,560 &  45,547,196 &  127,065 \\
\bottomrule
\end{tabular}
\end{table*}

Table \ref{tab:dataset-statistic} summarizes the statistics of the datasets used in our experiments. Among them, ogbg-code2, NA, and self-citation are directed acyclic graph (DAG) datasets, while MalNet-Tiny and EDA-HLS are directed cyclic graph datasets.

\textbf{DAG Datasets}:
\begin{itemize}
    \item \textbf{ogbg-code2} \cite{ogb-benchmark}: This dataset consists of Abstract Syntax Trees (ASTs) parsed from Python functions, with node features representing syntactic tokens. The graph-level classification task aims to predict the first five tokens of the function name.
    \item \textbf{NA} \cite{DVAE}: Contains neural network architectures generated by the ENAS framework, represented as directed acyclic graphs with a relatively small scale. Node features indicate the types of neural network components (e.g., convolutional layers, pooling layers), and the graph-level regression task aims to predict the performance of the architecture on the CIFAR-10 dataset.
    \item \textbf{self-citation} \cite{DAGFormer}: Each directed acyclic graph represents the self-citation network of a scholar. Node features include the publication year and the total number of citations (excluding the paper to be classified). The node-level classification task determines whether a paper belongs to the highly-cited category, serving as a proxy metric for academic influence.
\end{itemize}

\textbf{Directed Cyclic Graph Datasets}:
\begin{itemize}
    \item \textbf{MalNet-Tiny} \cite{malnet-dataset}: Contains 5,000 call graphs of malicious and benign software, covering 5 different categories. The graph-level classification task is to determine the category of each call graph.
    \item \textbf{EDA-HLS} \cite{EDA-hls-dataset}: Contains 18,750 intermediate representation (IR) graphs of C/C++ code processed through front-end compilation \cite{IR-graphs}. The dataset labels represent performance metrics after implementation on FPGA devices, obtained through synthesis using the Vitis HLS tool \cite{vitis-hls} combined with the Vivado \cite{vivado} implementation. The research task is to predict resource usage, specifically the utilization of Look-Up Tables (LUTs) and Digital Signal Processors (DSPs).
\end{itemize}

\subsubsection{Baselines}

We selected several representative baseline methods for comparison, which are broadly categorized into the following five groups.

\begin{itemize}
    \item \textbf{Undirected GNNs}: GCN \cite{GCN}, GAT \cite{GAT}, GIN \cite{GIN}
    \item \textbf{Directed GNNs}: DiGCN \cite{Digcn}, DirGNN \cite{dirgnn}, and Magnet \cite{magnet} designed for general directed graphs; DAGNN \cite{DAGNN} designed for DAGs
    \item \textbf{Undirected graph Transformers}: Graph Transformer (GT) \cite{GT}, SAT \cite{SAT} which employs a hybrid architecture of GNN and Transformer, GraphGPS \cite{GraphGPS} and Exphormer \cite{Exphormer} that adopt sparse attention mechanisms
    \item \textbf{Directed graph Transformers}: SAT+MagLapPe \cite{DirPE} utilizing magnetic Laplacian positional encoding for general directed graphs; DAG-specific Transformers including PACE \cite{PACE}, DAG+Transformer \cite{DAGFormer}, and DAG+SAT \cite{DAGFormer}
    \item \textbf{Graph Mamba Models}: Graph-Mamba \cite{GraphMamba} which sequentializes the entire graph into a node sequence; GMN \cite{GMN} that incorporates subgraphs during the sequentialization process
\end{itemize}

\subsubsection{Implementation details}

For baseline models, if the original papers have already reported experimental results on our selected datasets, we directly cite the results from those papers; otherwise, we perform hyperparameter tuning on the chosen datasets based on their officially recommended parameter settings and code implementations. For the DirGraphSSM model, we define the following hyperparameter search space:
\begin{itemize}

    \item \textbf{Model Architecture}: Number of DirGraphSSM layers $L \in \{1, 2, 3\}$, hidden dimension $d \in \{32, 64, 128\}$, number of attention heads $h \in \{2, 4, 8\}$, number of DirGatedGCN layers for structural encoding $L_{\text{SE}} \in \{1, 2\}$, hidden dimension of the SSM convolution kernel $d_{\text{ssm}} \in \{4, 8, 16\}$, and the range of discrete time step lengths used for initializing the SSM convolution kernel parameters: $\text{dt}_{\text{min}} \in \{5 \times 10^{-4}, 10^{-3}, 5 \times 10^{-3}\}$, $\text{dt}_{\text{max}} \in \{5 \times 10^{-2}, 10^{-1}, 5 \times 10^{-1}\}$;
    \item  \textbf{Optimizer}: AdamW \cite{AdamW} is used, with learning rate $lr$ $\in \{10^{-4}, 5 \times 10^{-4}, 10^{-3}, 5 \times 10^{-3}\}$, and weight decay $\lambda$ $\in \{10^{-6}, 5 \times 10^{-5}, 10^{-4}\}$;
    \item \textbf{Regularization}: Dropout rate $p$ $\in \{0.1, 0.3, 0.5\}$.
\end{itemize}

All experiments were conducted on a Linux server (1×Intel® Xeon® Gold 6240 CPU @ 2.60GHz, 1×NVIDIA A100 80GB PCIe, 263GB RAM).

For the ogbg-code2 and MalNet-Tiny datasets, we conducted systematic parameter analysis experiments on the maximum sequence length $K$ used for SSM scanning, as detailed in Section \ref{parameter-study-on-k}. For the EDA-HLS, Self-citation, and NA datasets, the value of $K$ was set heuristically based on their data statistics \ref{tab:dataset-statistic}: $K = 20$ for EDA-HLS, $K = 5$ for Self-citation, and $K = 7$ for NA.
On datasets of directed cyclic graphs, we replaced the DAG positional encoding in both DAG+Transformer and DAG+SAT with DepthPlus positional encoding, thereby eliminating the models’ dependency on acyclicity.

It is important to note that for the ogbg-code2 dataset, we computed the DepthPlus positional encoding using the original abstract syntax tree (AST) rather than the augmented graph structure generated in the official preprocessing step \cite{DAGFormer}. This is because the default graph augmentation process in this dataset introduces sequential connections between leaf nodes, causing the depth values of leaf nodes to be significantly larger than those of non-leaf nodes, thereby disrupting the inherent hierarchical relationships of the AST.

\subsection{Comparison on DAG Datasets}

\begin{table}[htbp]
\centering
\caption{Code graph classification results and avg training time on ogbg-code2. The top three models in terms of accuracy are highlighted with \textbf{bold}, \underline{underline} and \textit{italic} respectively.}
\label{tab:ogbg-code2-result}
\begin{tabular}{cccc}
\toprule
\multicolumn{1}{c}{Model} & \multicolumn{1}{c}{Test F1 (\%) ↑} & \multicolumn{1}{c}{Valid F1 (\%) ↑}  & Time (epoch) \\
\midrule
GCN & 15.1 ± 0.2 & 14.0 ± 0.2 & 249s  \\
GAT & 15.7 ± 0.2 & 14.4 ± 0.2 & 265s  \\
GIN & 14.9 ± 0.2 & 13.7 ± 0.2 & 349s  \\
\midrule
DiGCN & 16.7 ± 0.2 & 15.6 ± 0.2 & 474s  \\
DirGNN & 16.6 ± 0.2 & 15.5 ± 0.2 & 432s  \\
Magnet & 16.9 ± 0.2 & 15.7 ± 0.2 & 465s  \\
DAGNN & 17.5 ± 0.5 & 16.1 ± 0.4 & 6018s \\
\midrule
PACE & 17.8 ± 0.2 & 16.3 ± 0.3 & 4365s  \\
SAT & 19.4 ± 0.3 & 17.7 ± 0.2 & 4408s  \\
SAT+maglap & - & - & OOT \\
GraphGPS & 18.9 ± 0.2 & 17.4 ± 0.2 & 3474s  \\
Exphormer & 19.9 ± 0.2 & 18.2 ± 0.1 & 2123s  \\
DAG+Transformer & 18.8 ± 0.2 & 17.4 ± 0.1 & 626s   \\
DAG+SAT & 20.2 ± 0.2 & 18.5 ± 0.1 & 871s  \\
\midrule
GMN & \textbf{20.6±0.20} & \textbf{18.9±0.13} & 1958s   \\
Graph-Mamba & \textit{20.3±0.23} & \textit{18.6±0.16} & 1836s   \\
DirGraphSSM & \underline{20.5±0.23} & \underline{18.7±0.15}  & 959s   \\
\bottomrule
\end{tabular}
\end{table}

\begin{table}[htbp]
\centering
\caption{Predictive performance of latent representations over NA.}
\label{tab:NA-result}
\begin{tabular}{ccc}
\toprule
\multicolumn{1}{c}{Model} & \multicolumn{1}{c}{RMSE ↓} & \multicolumn{1}{c}{Pearson’s r ↑}    \\
\midrule
GCN & 0.482 ± 0.003 & 0.871 ± 0.001   \\
GAT & 0.404 ± 0.003 & 0.894 ± 0.001   \\
GIN & 0.378 ± 0.003 & 0.905 ± 0.001   \\
\midrule
DiGCN & 0.269 ± 0.002 & 0.962 ± 0.001   \\
DirGNN & 0.273 ± 0.003 & 0.961 ± 0.001   \\
Magnet & 0.270 ± 0.002 & 0.961 ± 0.001    \\
DAGNN & 0.264 ± 0.004 & 0.964 ± 0.001  \\
\midrule
PACE & 0.254 ± 0.002 & 0.964 ± 0.001   \\
GT(GraphTransformer) & 0.275 ± 0.003 & 0.961 ± 0.001   \\
SAT & 0.298 ± 0.003 & 0.952 ± 0.001   \\
SAT+maglap & 0.269 ± 0.003 & 0.962 ± 0.001  \\
GraphGPS & 0.306 ± 0.004 & 0.950 ± 0.001   \\
Exphormer & 0.283 ± 0.003 & 0.953 ± 0.001    \\
DAG+Transformer & \textit{0.253 ± 0.002} & \textit{0.966 ± 0.001}   \\
DAG+SAT & 0.262 ± 0.004 & 0.964 ± 0.001    \\
\midrule
GMN & \underline{0.251 ± 0.003}  & \underline{0.967 ± 0.001}   \\
Graph-Mamba & 0.254 ± 0.002 & 0.965± 0.001   \\
DirGraphSSM & \textbf{0.249 ± 0.002} & \textbf{0.969 ± 0.001}     \\
\bottomrule
\end{tabular}
\end{table}

\begin{table}[htbp]
\centering
\caption{Node classification results for the self-citation dataset. }
\label{tab:self-citation-result}
\begin{tabular}{ccc}
\toprule
\multicolumn{1}{c}{Model} & \multicolumn{1}{c}{AP (\%) ↑} & \multicolumn{1}{c}{ROC-AUC (\%) ↑}    \\
\midrule
GCN & 58.8 ± 0.4 & 79.9 ± 0.2   \\
GAT & 55.3 ± 3.7 & 77.9 ± 1.4  \\
GIN & 57.7 ± 1.8 & 79.7 ± 0.2   \\
\midrule
DiGCN & 62.6 ± 0.9 & 80.3 ± 0.8   \\
DirGNN & 62.3 ± 0.7 & 80.1 ± 0.8   \\
Magnet & 62.5 ± 1.1 & 80.4 ± 0.8    \\
DAGNN & 61.2 ± 0.6 & 81.0 ± 0.3  \\
\midrule
SAT & 59.8 ± 1.7 & 79.8 ± 0.7   \\
SAT+maglap & 62.7 ± 2.5 & 81.8 ± 0.6  \\
GraphGPS & 61.6 ± 2.6 & 81.3 ± 0.6   \\
Exphormer & 62.4 ± 2.2 &  81.5 ± 0.5    \\
DAG+Transformer & 63.8 ± 0.8 & 82.2 ± 0.5  \\
DAG+SAT & 62.7 ± 1.5 & 80.6 ± 0.7   \\
\midrule
GMN & \underline{65.4 ± 1.3}  & \underline{82.7 ± 0.6}  \\
Graph-Mamba & \textit{64.7 ± 1.1} & \textit{82.5 ± 0.4}  \\
DirGraphSSM & \textbf{65.9 ± 1.6} & \textbf{82.8 ± 0.7}     \\
\bottomrule
\end{tabular}
\end{table}

Many real-world problems can be modeled as directed acyclic graphs (DAGs). Due to the acyclic nature, DAGs exhibit a clear hierarchical partitioning (i.e., partial order) among all nodes connected by  directed  reachable paths, resulting in typical long-range causal dependencies that significantly impact downstream tasks.

We conducted 10 experiments with different random seeds on selected DAG datasets and recorded the mean and standard deviation of each evaluation metric. Detailed results are shown in Table \ref{tab:ogbg-code2-result}, Table \ref{tab:NA-result}, and Table \ref{tab:self-citation-result} . 
Overall, the results from various baseline models indicate that directed GNNs outperform undirected  GNNs, and the best directed graph Transformer also surpasses the best undirected graph Transformer, highlighting the necessity of explicitly modeling edge directionality in directed graph learning. Meanwhile, graph Mamba models outperform the best graph Transformer, demonstrating the effectiveness of the state space model (SSM) approach.

From the perspective of DirGraphSSM, on small-scale datasets including NA and selfCitation, our proposed DirGraphSSM achieves state-of-the-art (SOTA) accuracy. On the very large-scale dataset ogbg-code2, it attains the second-highest accuracy while requiring only 49\% of the training time per epoch compared to the current SOTA method GMN, thus achieving a favorable balance between accuracy and efficiency.
From the architectural design perspective of DirGraphSSM, we further compare and analyze its overall experimental performance against various baselines:
\begin{itemize}    
    \item  Compared with directed GNNs, our model captures long-range causal dependencies and node hierarchy through DirEgo2Token and DepthPlus positional encoding, leading to higher accuracy;
    \item  Compared with graph Transformers, our model restricts the receptive field of message passing via DirEgo2Token’s unidirectional neighborhood expansion, reducing interference from irrelevant noise. Moreover, by effectively integrating the SSM mechanism with the attention mechanism, it achieves superior modeling of node feature and topological structure, significantly surpassing the best graph Transformer in both accuracy and efficiency;
    \item  Compared with sequence scanning–based graph mamba models, our model achieves fully parallel processing via a message passing–based SSM scanning mechanism. Combined with lightweight structural encoding, positional encoding, and Digraph Fusion Attention modules, It achieves higher efficiency while maintaining accuracy.
\end{itemize}

\subsection{Comparison on Directed Cyclic Graph Datasets}

\begin{table}[htbp]
\centering
\caption{statistic of Directed Cyclic Graph Datasets.}
\label{tab:statistic-directed-cyclic-dataset}
\begin{tabular}{ccc}
\toprule
\multicolumn{1}{c}{cycle statistic} & \multicolumn{1}{c}{MalNet-Tiny} & \multicolumn{1}{c}{HLS}    \\
\midrule
Num Graphs & 5,000 & 18,570   \\
Avg Nodes per Graph & 1,410.30 & 95   \\
\midrule
Avg Cycle Nodes per Graph & 4.69 & 10.45   \\
Avg Cycle Nums per Graph &  1.33  & 0.85   \\
Avg Cycle Size & 3.53 & 12.35  \\
\bottomrule
\end{tabular}
\end{table}

\begin{table}[htbp]
\centering
\caption{graph classification accuracy and avg training time for the MalNet-Tiny dataset. The top three models in terms of accuracy are highlighted with \textbf{bold}, \underline{underline} and \textit{italic} respectively.}
\label{tab:MalNet-Tiny-result}
\begin{tabular}{ccc}
\toprule
\multicolumn{1}{c}{Model} & \multicolumn{1}{c}{Accuracy (\%) ↑} & \multicolumn{1}{c}{Time (epoch)}    \\
\midrule
GCN & 81.00±0.01 & 12s   \\
GAT & 85.09±0.25 & 13s   \\
GIN & 88.98±0.55 & 16s   \\
\midrule
DiGCN & 90.81±0.17 & 22s   \\
DirGNN & 90.73±0.15 & 20s   \\
Magnet & 90.95±0.25 & 23s     \\
DAGNN & - & -  \\
\midrule
PACE & - & -  \\
SAT & 92.56±0.31 & 206s   \\
SAT+maglap & - & OOT  \\
GraphGPS & 92.64±0.78 & 163s   \\
Exphormer & 94.02±0.20 &  102s    \\
DAG+Transformer-plus & 94.04±0.19 & 35s   \\
DAG+SAT-plus & \textit{94.07±0.20} & 48s    \\
\midrule
GMN & \textbf{94.15±0.20}  & 91s   \\
Graph-Mamba & 94.06±0.18 & 86s   \\
DirGraphSSM & \underline{94.09±0.23} & 61s     \\
\bottomrule
\end{tabular}
\end{table}

\begin{table}[htbp]
\centering
\caption{graph regression results for the EDA-HLS dataset.}
\label{tab:EDA-HLS-result}
\begin{tabular}{ccc}
\toprule
\multicolumn{1}{c}{Model} & \multicolumn{1}{c}{MSE-DSP ↓} & \multicolumn{1}{c}{MSE-LUT ↓}    \\
\midrule
GCN & 2.903±0.083 & 2.305±0.079   \\
GAT & 2.865±0.085 & 2.293±0.089   \\
GIN & 2.813±0.091 & 2.287±0.084   \\
\midrule
DiGCN & 2.717±0.094 & 2.124±0.078  \\
DirGNN & 2.526±0.147 & 2.017±0.096   \\
Magnet & 2.637±0.091 & 2.084±0.098    \\
DAGNN & - & -  \\
\midrule
PACE & - & -  \\
SAT & 2.631±0.106 & 2.119±0.127  \\
SAT+maglap & 2.453±0.115 & 2.139±0.101  \\
GraphGPS & 2.615±0.106 & 2.136±0.124  \\
Exphormer & 2.517±0.099 &  2.009±0.115    \\
DAG+Transformer-plus & 2.369±0.075 & 1.946±0.091   \\
DAG+SAT-plus & 2.327±0.079 & 1.894±0.095    \\
\midrule
GMN & \underline{2.267±0.126}  & \underline{1.795±0.115}  \\
Graph-Mamba & \textit{2.302±0.118} & \textit{1.837±0.104}   \\
DirGraphSSM & \textbf{2.203±0.124} & \textbf{1.702±0.107}     \\
\bottomrule
\end{tabular}
\end{table}

In real-world directed graphs, cycles (where cycles refer to strongly connected components with more than one node) are often present. Nodes within these cycles cannot be strictly ordered in a hierarchical manner, yet the acyclic parts of the graph may still exhibit significant hierarchical structure and causal dependencies. According to the statistical results in Table \ref{tab:statistic-directed-cyclic-dataset}, although directed graphs in the MalNet-Tiny and EDA-HLS datasets contain cycles, the average number of nodes within these cycles is much smaller than the total number of nodes in the graph, indicating that most nodes are not part of any cycle. Therefore, these directed graphs still have a clear hierarchy structure, making them relatively similar to Directed Acyclic Graphs (DAGs). Learning long-range causal dependencies remains an important task in such directed cyclic graphs that have hierarchical structures.

We conducted 10 independent experiments with different random seeds on the above directed cyclic graph datasets and recorded the mean and standard deviation of each evaluation metric. The detailed results are shown in Table \ref{tab:MalNet-Tiny-result} and Table \ref{tab:EDA-HLS-result}. Experimental results demonstrate that our proposed DirGraphSSM model achieves state-of-the-art (SOTA) performance on the medium-scale EDA-HLS dataset. On the large-scale MalNet-Tiny dataset, it also achieves the second-best performance, while the average training time per epoch is only 66\% that of the current SOTA method, GMN, thus striking a good balance between accuracy and efficiency.

Overall, the trends observed for various baseline models and our proposed DirGraphSSM on these directed cyclic graph datasets are largely consistent with the conclusions drawn on DAG datasets. In particular, the DepthPlus method adopted by DirGraphSSM uses Tarjan’s algorithm \cite{Tarjan-algorithm} to recover the hierarchical structure in directed cyclic graphs, overcoming the acyclicity constraints of existing directed GNNs designed for DAGs. Leveraging this hierarchical information, DirGraphSSM achieves excellent accuracy performance.

\subsection{Comparison on efficiency}

Our DirGraphSSM model employs a parallel message-passing mechanism, enabling an efficient SSM scanning process and avoiding additional padding operations, thereby demonstrating significant computational efficiency advantages. This is further validated by the training time statistics on the very large-scale dataset ogbg-code2 and the large-scale dataset MalNet-Tiny. As shown in Table \ref{tab:ogbg-code2-result} and Table \ref{tab:MalNet-Tiny-result} , DirGraphSSM not only outperforms mainstream graph Transformer models in training speed but is also significantly faster than existing Graph Mamba models. Visual comparisons of training times are provided in Fig. \ref{fig:k-MalNet-Tiny} and Fig. \ref{fig:k-ogbg-code2}.

The specific comparison results are as follows:  
\begin{itemize}
    \item \textbf{Compared with Graph Mamba-style models}:  
        \begin{itemize}
            \item  On the ogbg-code2 dataset, DirGraphSSM’s average training time per epoch is only 0.49× that of GMN and 0.52× that of Graph-Mamba.  
            \item  On the MalNet-Tiny dataset, DirGraphSSM’s average training time per epoch is 0.67× that of GMN and 0.71× that of Graph-Mamba.  
        \end{itemize}
    \item \textbf{Compared with graph Transformer models} :
        \begin{itemize}
            \item On the ogbg-code2 dataset, DirGraphSSM’s average training time per epoch is 0.28× that of GraphGPS, 0.45× that of Exphormer, and 0.21× that of SAT.
            \item On the MalNet-Tiny dataset, DirGraphSSM’s average training time per epoch is 0.37× that of GraphGPS, 0.6× that of Exphormer, and 0.29× that of SAT.  
        \end{itemize}
\end{itemize}

Compared to DAGFormer models (e.g., DAG+Transformer and DAG+SAT) \cite{DAGFormer}, which utilize message passing and reachability attention mechanisms, DirGraphSSM requires processing with an SSM convolutional kernel during the message-passing phase, resulting in slightly slower training speed. However, thanks to the superior long-range dependency modeling capability of the SSM kernel, DirGraphSSM achieves higher accuracy with relatively low computational overhead, significantly outperforming DAGFormer-style models in performance.

Furthermore, the SAT model based on magnetic Laplacian positional encodings encountered Out-of-Time (OOT) issues on the two large-scale datasets mentioned above. This is due to the high computational complexity of such positional encodings, which rely on matrix factorization and eigenvector basis transformation, leading to severe scalability bottlenecks on large graph datasets. This observation also highlights the importance of adopting lightweight positional encoding strategies for large-scale graph data.

\subsection{ablation study}

\begin{table*}[t]
\centering
\caption{Ablation Results.}
\label{tab:ablation}
\begin{tabular}{@{}lccccccccc@{}}
\toprule
\multirow{2}{*}{Ablation module} & 
\multicolumn{1}{c}{ogbg-code2} & 
\multicolumn{1}{c}{MalNet-Tiny} & 
\multicolumn{2}{c}{NA} & 
\multicolumn{2}{c}{Self-citation} & 
\multicolumn{2}{c}{EDA-HLS} \\
\cmidrule(lr){2-2} \cmidrule(lr){3-3} \cmidrule(lr){4-5} \cmidrule(lr){6-7} \cmidrule(lr){8-9}
& Test F1 \% & Accuracy \% & RMSE $\downarrow$ & Pearson’s $r$ $\uparrow$ & AP $\uparrow$ & ROC-AUC $\uparrow$ & MSE-DSP $\downarrow$ & MSE-LUT $\downarrow$ \\
\midrule
DirGraphSSM & 20.5$\pm$0.23 & 94.09$\pm$0.23 & 0.249$\pm$0.002 & 0.969$\pm$0.001 & 65.9$\pm$1.6 & 82.8$\pm$0.7 & 2.203$\pm$0.124 & 1.702$\pm$0.107 \\
(-) DepthPlus-PE & 20.1$\pm$0.21 & 93.79$\pm$0.25 & 0.260$\pm$0.002 & 0.961$\pm$0.001 & 64.9$\pm$1.4 & 82.5$\pm$0.5 & 2.253$\pm$0.118 & 1.786$\pm$0.104 \\
(-) DirGatedGCN-SE & 19.9$\pm$0.19 & 93.37$\pm$0.19 & 0.251$\pm$0.001 & 0.967$\pm$0.001 & 65.2$\pm$1.2 & 82.6$\pm$0.8 & 2.276$\pm$0.126 & 1.753$\pm$0.109 \\
(-) DigraphFusionAttention & 20.3$\pm$0.17 & 93.98$\pm$0.21 & 0.256$\pm$0.002 & 0.964$\pm$0.001 & 65.4$\pm$1.7 & 82.4$\pm$0.4 & 2.225$\pm$0.127 & 1.731$\pm$0.113 \\
(-) SSM Scan Kernel & 19.4$\pm$0.18 & 93.02$\pm$0.18 & 0.264$\pm$0.002 & 0.960$\pm$0.001 & 64.6$\pm$1.5 & 82.2$\pm$0.6 & 2.314$\pm$0.122 & 1.868$\pm$0.105 \\
\bottomrule
\end{tabular}
\end{table*}

To validate the effectiveness of the four core components (DepthPlus-PE, DirGatedGCN-SE, Digraph Fusion Attention, and SSM Scan Kernel) in our  proposed DirGraphSSM , we conducted a series of ablation studies across all datasets. The experimental results are shown in Table \ref{tab:ablation}.

Overall, removing any of the core modules leads to a decline in model performance. This phenomenon is consistently observed across different datasets, indicating that the proposed lightweight positional encoding, structural encoding, SSM scanning mechanism, and Digraph Fusion Attention module are all beneficial for directed graph learning. In particular, the most pronounced performance drop occurs when the SSM Scan Kernel is removed, demonstrating that incorporating the SSM scanning mechanism effectively captures long-range causal dependencies in directed graphs, thereby significantly improving model performance.

\subsection{Parameter Study on SSM Seq Length}
\label{parameter-study-on-k}


\begin{figure*}[t]
    
    \begin{minipage}{0.48\textwidth}
        \centering
        \captionof{table}{Parameter Study of ssm seq len $k$ on ogbg-code2.}
        \label{tab:k-ogbg-code2}
        \begin{tabular}{cccc}
            \toprule
            \multicolumn{1}{c}{$k$} & \multicolumn{1}{c}{Avg $p_k$} & \multicolumn{1}{c}{Time (epoch)}  & \multicolumn{1}{c}{Test F1(\%)}  \\
            \midrule
            1 & 0.99  & 780.9s & 15.56±0.96   \\
            2 & 1.96  & 824.6s & 19.46±0.11   \\
            3 & 2.85  & 864.7s & 19.63±0.12   \\
            4 &  3.58   & 897.9s & 20.04±0.17   \\
            5 & 4.15  & 923.4s & 20.36±0.06  \\
            6 & 4.51   & 939.6s & 20.37±0.14  \\
            7 & 4.70  & 948.4s & 20.38±0.13 \\
            8 &  4.80  & 952.9s & 20.4±0.11 \\
            9 & 4.85  & 958.7s & 20.51±0.23 \\
            \bottomrule
        \end{tabular}
    \end{minipage}
    \hfill
    \begin{minipage}{0.48\textwidth}
        \centering
        \captionof{table}{Parameter Study of ssm seq len $k$ on MalNet-Tiny.}
        \label{tab:k-MalNet-Tiny}
        \begin{tabular}{cccc}
            \toprule
            \multicolumn{1}{c}{$k$} & \multicolumn{1}{c}{Avg $p_k$} & \multicolumn{1}{c}{Time (epoch)}  & \multicolumn{1}{c}{Test F1(\%)}  \\
            \midrule
            1 & 1.88  & 42.2s & 90.59±0.16   \\
            2 & 3.63  & 46.9s & 91.42±0.13   \\
            3 & 5.17  & 51.0s & 92.35±0.14   \\
            4 &  6.32   & 54.1s & 92.9±0.14   \\
            5 & 7.15  & 56.4s & 93.64±0.18  \\
            6 & 7.72   & 57.9s & 93.83±0.2  \\
            7 & 8.09  & 58.9s & 93.97±0.22 \\
            8 &  8.33  & 59.6s & 94.03±0.21 \\
            9 & 8.62  & 60.8s & 94.09±0.23 \\
            \bottomrule
        \end{tabular}
    \end{minipage}
    
\end{figure*}


\begin{figure*}[t]
    \begin{minipage}{0.48\textwidth}
        \centering
        \includegraphics[width=\linewidth, height=0.8\textheight, keepaspectratio]{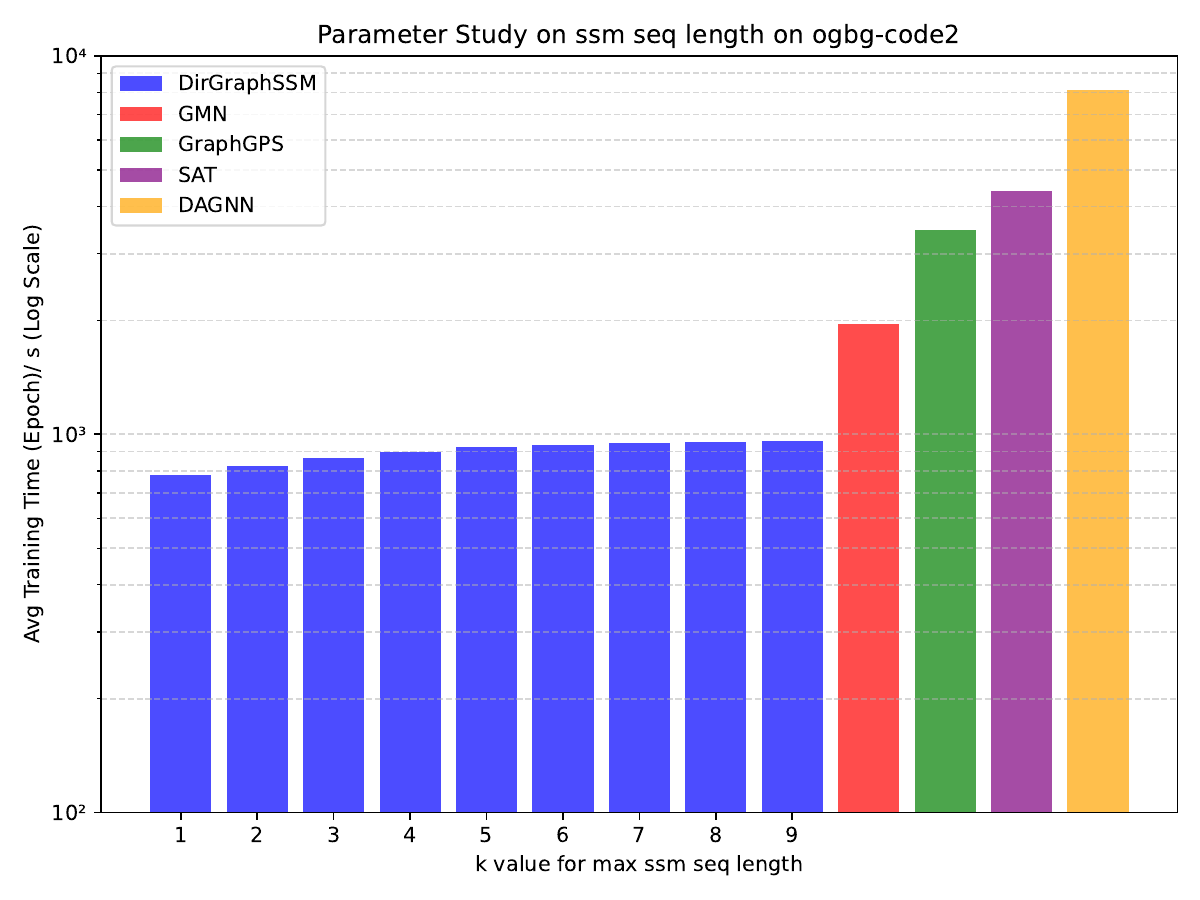}
        \captionof{figure}{Average training time per epoch for various ssm seq length K over ogbg-code2, log scale.}
        \label{fig:k-ogbg-code2}
    \end{minipage}
    \hfill
    \begin{minipage}{0.48\textwidth}
        \centering
        \includegraphics[width=\linewidth, height=0.8\textheight, keepaspectratio]{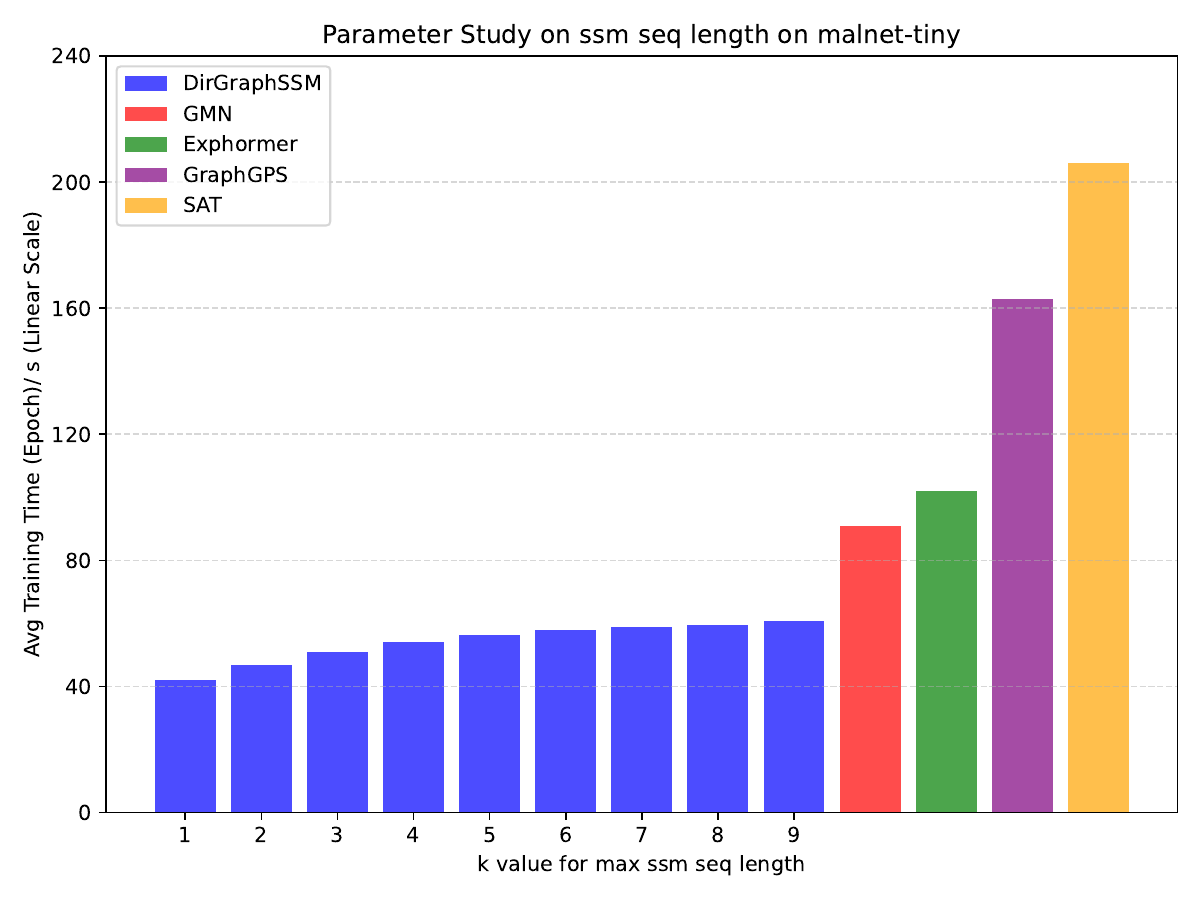}
        \captionof{figure}{Average training time per epoch for various ssm seq length K over MalNet-Tiny, linear scale.}
        \label{fig:k-MalNet-Tiny}
    \end{minipage}
    
\end{figure*}


To investigate the impact of the SSM sequence length on model performance and efficiency, we conducted an empirical study on the maximum SSM sequence length $K$ using the ogbg-code2 and MalNet-Tiny datasets. The corresponding results are shown in Table \ref{tab:k-ogbg-code2} and Table \ref{tab:k-MalNet-Tiny}.

First, we examine the influence of the maximum SSM sequence length $K$ on the number of message passing steps. Traditional GNNs typically limit message passing to 2 or 3 hops, resulting in a limited receptive field. Therefore, using an infinite number of hops ($K = \infty$) may seem impractical. However, in real-world directed graphs, if we only consider nodes along unidirectionally reachable paths, the number of nodes within the receptive field when $K = \infty$ is significantly reduced and is often much smaller than the total number of nodes $N$ in the graph. As can be seen from Table \ref{tab:dataset-statistic}, in the ogbg-code2, MalNet-Tiny, and self-citation datasets, the average number of predecessors per node ($p_\infty$) is much smaller than the average number of nodes in the graph, making the number of message passing steps approximately linear with respect to the number of nodes. On the NA dataset, $p_\infty$ is about $1/2$ of the total number of nodes, while on the EDA-HLS dataset it is about $1/4$, leading to a relatively higher number of message passing steps. Fortunately, both of these datasets have small graph sizes and a limited number of graphs, so the overall $\text{Total}_{p_\infty}$ is significantly smaller than that of large-scale datasets such as ogbg-code2 and MalNet-Tiny. Thus, the computational complexity remains within an acceptable range.

Furthermore, we analyze the effect of different maximum SSM sequence lengths $K$ on model training time and performance. Experimental results (see Table \ref{tab:k-ogbg-code2} and Table \ref{tab:k-MalNet-Tiny}) show that as $K$ increases, the model achieves a slight but consistent improvement in accuracy by incorporating longer-range causal information, while the training time increases marginally. This phenomenon indicates the necessity of learning long-range causal relationships in directed graphs and also suggests that the value of $K$ can be adjusted to balance efficiency and accuracy (see Figure \ref{fig:k-MalNet-Tiny} and Figure \ref{fig:k-ogbg-code2}).

\section{CONCLUSION}
In this paper, we first propose DirGraphSSM, a novel graph state space model designed for large-scale sparse directed graph learning. Through two innovative components, namely \textbf{DirEgo2Token} and \textbf{Digraph SSM Scan}. Digraph SSM Scan module efficiently performs SSM scanning on large sparse directed graphs using a message-passing mechanism, thereby effectively capturing long-range causal dependencies. To overcome the performance degradation caused by the time invariance of traditional SSM convolutional kernels while maintaining computational efficiency, we innovatively introduce a multi-head attention mechanism into SSM scanning process, achieving an effect similar to the selective scanning in Mamba models without compromising parallelism. To further enhance the model's awareness of directed graph topology, we design DepthPlus positional encoding, DirGatedGCN structural encoding, and a Digraph Fusion Attention module, which strengthen the model’s representational capacity from three dimensions: global, local, and multi-head attention spaces. Experiments on multiple   datasets demonstrate that DirGraphSSM outperforms representative directed GNNs and graph transformers, and achieves a better efficiency-accuracy trade-off compared to state-of-the-art graph state space models (e.g.,GMN and Graph Mamba).

Although DirGraphSSM can efficiently process large batches of sparse directed graphs, it suffers from significant performance degradation when handling dense directed graphs due to the sequentialization of K-hop directed neighborhoods in DirEgo2Token. Moreover, how to choose the maximum sequence length K for the SSM to balance efficiency and accuracy remains an open question. Therefore, critical future work includes designing new strategies (e.g., neighborhood sampling strategies) to address the limitations on dense directed graphs and developing more effective methods for selecting the maximum sequence length K in SSMs.


\bibliographystyle{IEEEtran}
\bibliography{IEEEfull, DirGraphSSM}



 
\vspace{11pt}


\begin{IEEEbiography}[{\includegraphics[width=1in,height=1.25in,clip,keepaspectratio]{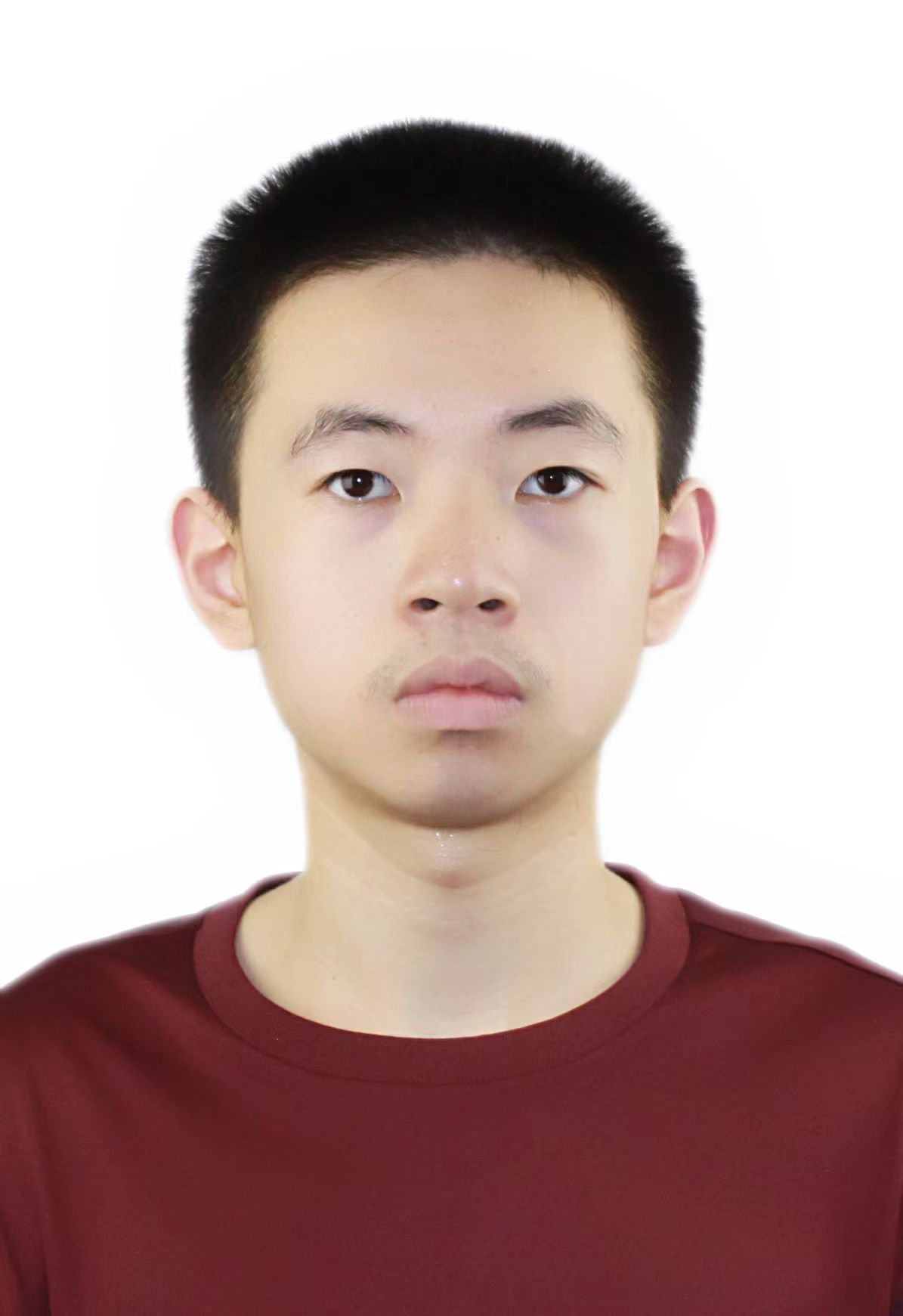}}]{Junzhi She}
Junzhi She is currently working toward the Master's degree with the school of Computer Science, Beijing Institute of Technology, advised by Prof. Guoren Wang. He received the BS degree in computer science from Beijing Institute of Technology in 2024. His research interest lies in graph machine learning and AI Agent systems.
\end{IEEEbiography}

\vspace{11pt}

\begin{IEEEbiography}[{\includegraphics[width=1in,height=1.25in,clip,keepaspectratio]{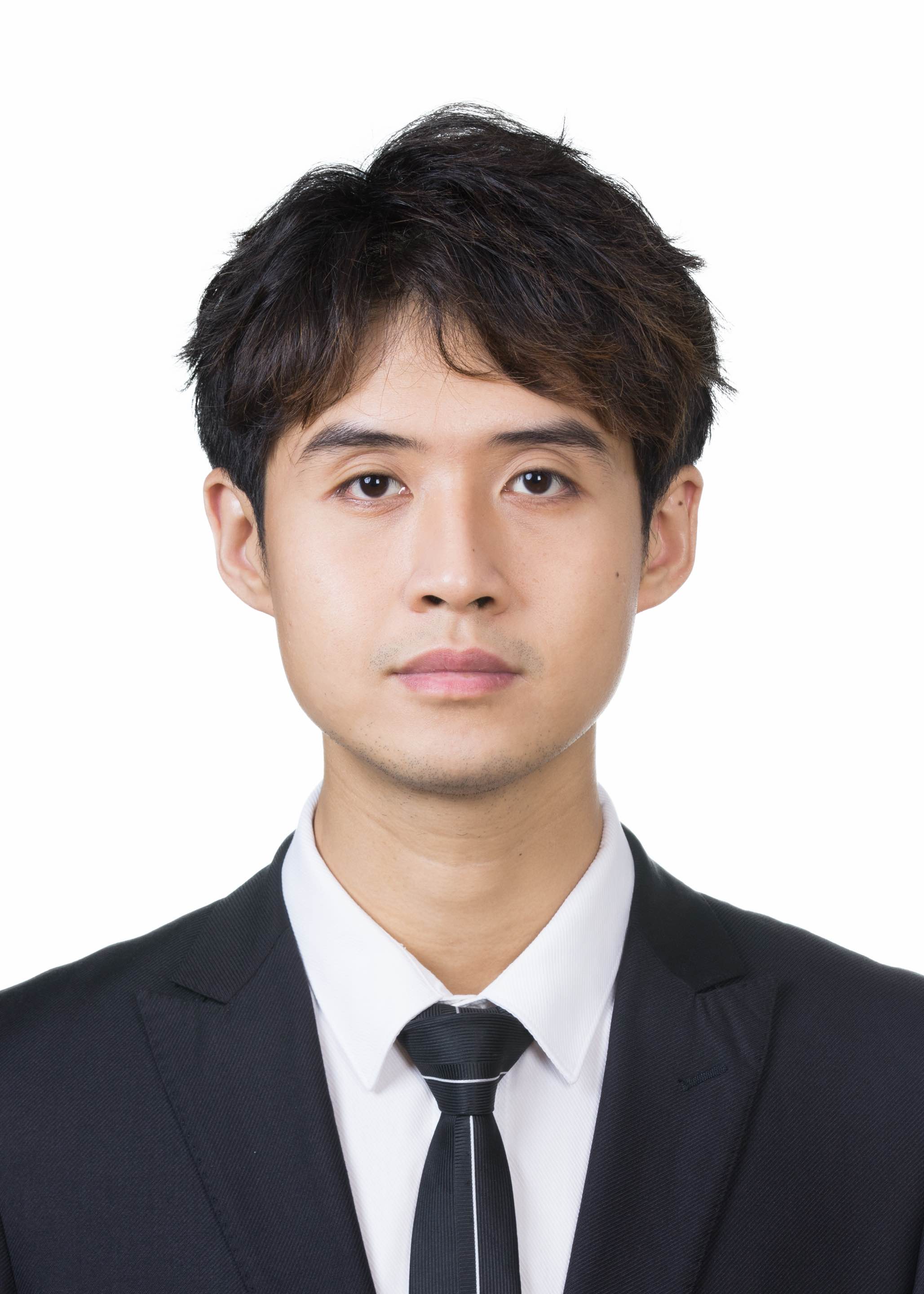}}]{Xunkai Li}
Xunkai Li is currently working toward the PhD degree with the school of Computer Science, Beijing Institute of Technology, advised by Prof. Rong-hua Li. He received the BS degree in computer science from Shandong University in 2022. His research interest lies in Data-centric Graph Intelligence (Data-centric AI, Graph Machine Learning, and AI4Science). He has published 10+ papers in top ML/DB/DM/AI conferences such as ICML, VLDB, WWW, AAAI.
\end{IEEEbiography}

\vspace{11pt}

\begin{IEEEbiography}[{\includegraphics[width=1in,height=1.25in,clip,keepaspectratio]{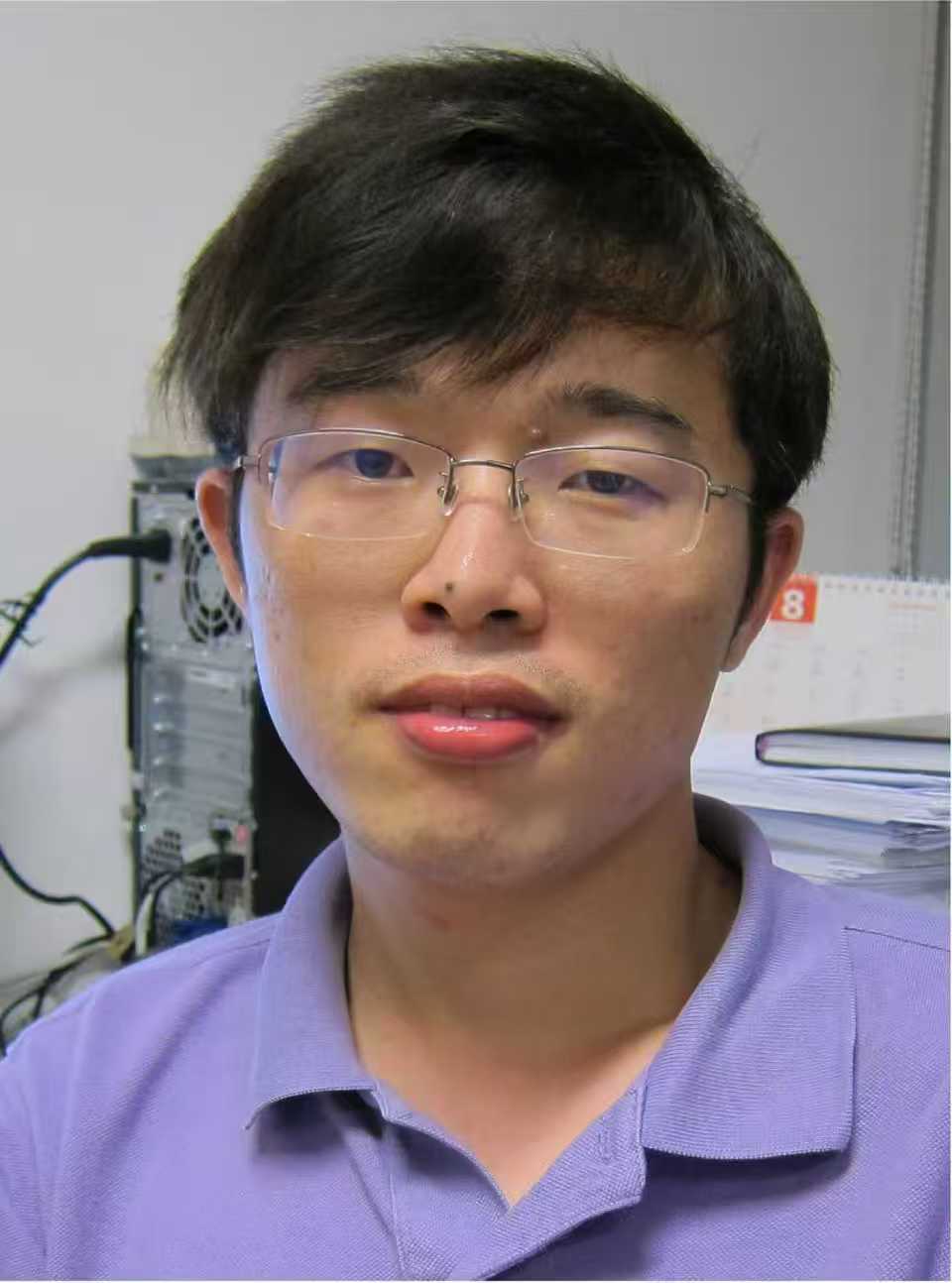}}]{Rong-Hua Li}
Rong-Hua Li received the PhD degree from the Chinese University of Hong Kong, in 2013. He is currently a professor with the Beijing Institute of Technology (BIT), Beijing, China. His research interests include graph data management and mining, social network analysis, graph computational systems, and graph-based machine learning.
\end{IEEEbiography}

\vspace{11pt}

\begin{IEEEbiography}[{\includegraphics[width=1in,height=1.25in,clip,keepaspectratio]{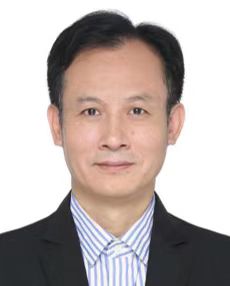}}]{Guoren Wang}
Guoren Wang received the BS, MS, and PhD degrees from the Department of Computer Science, Northeastern University, China, in 1988, 1991, and 1996, respectively. Currently, he is a professor at the Beijing Institute of Technology (BIT), Beijing, China. His research interests include graph data management, graph mining, and graph computational systems.
\end{IEEEbiography}

\vfill

\end{document}